\documentclass[10pt,twocolumn,letterpaper]{article}

\usepackage{wacv}              %

\usepackage{graphicx}
\usepackage{amsmath}
\usepackage{amssymb}
\usepackage{booktabs}
\usepackage[symbol]{footmisc}

\usepackage[ruled,linesnumbered]{algorithm2e}
\usepackage[accsupp]{axessibility}  %

\usepackage[pagebackref,breaklinks,colorlinks]{hyperref}

\hypersetup{
    linkcolor=magenta      %
}

\usepackage[capitalize]{cleveref}
\crefname{section}{Sec.}{Secs.}
\Crefname{section}{Section}{Sections}
\Crefname{table}{Table}{Tables}
\crefname{table}{Tab.}{Tabs.}

\usepackage{cuted}
\usepackage{caption}

\usepackage{enumitem}

\DeclareMathOperator{\G}{\mathbf{G}} %
\DeclareMathOperator{\adaattn}{\mathbf{M}} %
\DeclareMathOperator{\concat}{aggregate}
\DeclareMathOperator*{\argmin}{arg\,min}
\newcommand{\isi}{\{I_i\}_{i=1}^{N_I}} %
\newcommand{\ist}{\{T_i\}_{i=1}^{N_T}} %
\newcommand{\sr}{\{\mathcal{S}_i\}_{i=1}^{N_\mathcal{S}}} %

\newcommand{\ii}{I_i}
\newcommand{\ti}{T_i}
\newcommand{\si}{\mathcal{S}}

\newcommand{\et}{E_T}
\newcommand{\ei}{E_I}

\DeclareMathOperator{\aug}{aug}
\DeclareMathOperator{\crop}{crop}

\def\wacvPaperID{362} %
\def\confName{WACV}
\def\confYear{2024}

\begin{document}

\title{Multimodality-guided Image Style Transfer using Cross-modal GAN Inversion}

\author{\normalsize
Hanyu Wang$^1$\footnote[2]{}, Pengxiang Wu$^2$, Kevin Dela Rosa$^2$, Chen Wang$^2$, Abhinav Shrivastava$^1$ \\
\normalsize
$^1$University of Maryland, College Park \qquad $^2$Snap Inc.\\
{\tt \footnotesize hywang66@umd.edu
\{pwu,kevin.delarosa,chen.wang\}@snapchat.com} {\tt \footnotesize abhinav@cs.umd.edu}
}

\maketitle

\footnotetext[2]{Work done during internship at Snap Inc.}

\begin{strip}
    \centering
    \vspace{-0.3cm}%
    \includegraphics[width=\linewidth]{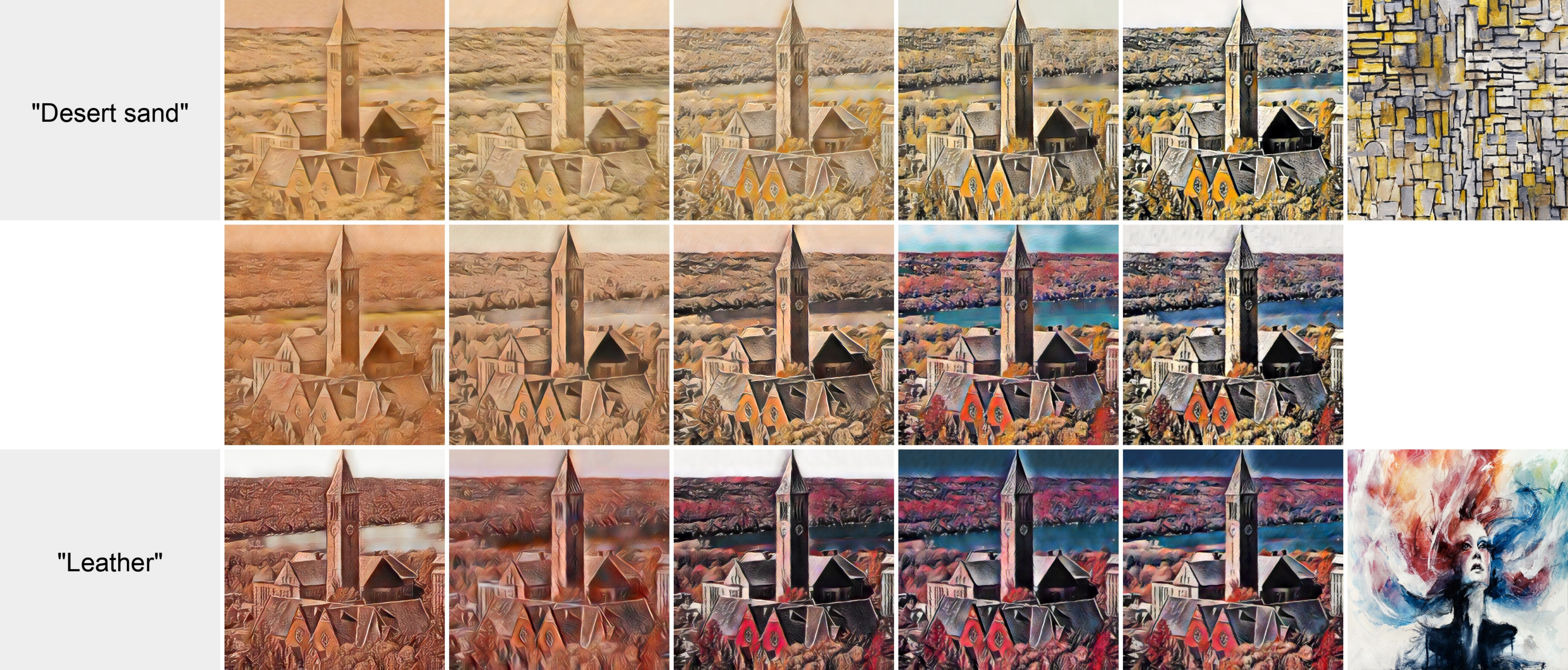}
    \captionof{figure}{\textbf{Our multimodality-guided image style transfer results.} 
    We show style transfer results guided by two text styles (left) and two imag styles (right). 15 stylized images are synthesized by evenly interpolating between these four styles.
    }
    \label{fig:teaser}
\end{strip}

\begin{abstract}
    Image Style Transfer (IST) is an interdisciplinary topic of computer vision and art that continuously attracts researchers’ interests. Different from traditional Image-guided Image Style Transfer (IIST) methods that require a style reference image as input to define the desired style, recent works start to tackle the problem in a text-guided manner, \ie, Text-guided Image Style Transfer (TIST). Compared to IIST, such approaches provide more flexibility with text-specified styles, which are useful in scenarios where the style is hard to define with reference images. 
    Unfortunately, many TIST approaches produce undesirable artifacts in the transferred images.
    To address this issue, we present a novel method to achieve much improved style transfer based on text guidance. Meanwhile, to offer more flexibility than IIST and TIST, our method allows style inputs from multiple sources and modalities, enabling MultiModality-guided Image Style Transfer (MMIST).
    Specifically, we realize MMIST with a novel cross-modal GAN inversion method, which generates style representations consistent with specified styles. Such style representations facilitate style transfer and in principle generalize any IIST methods to MMIST.
    Large-scale experiments and user studies 
    demonstrate that our method achieves state-of-the-art performance on TIST task. Furthermore, comprehensive qualitative results confirm the effectiveness of our method on MMIST task and cross-modal style interpolation. 

    \noindent Project website: \href{https://hywang66.github.io/mmist}{https://hywang66.github.io/mmist}.
\end{abstract}

\section{Introduction}
\label{sec:intro}

As a research topic at the intersection of computer vision and art, Image Style Transfer (IST) aims to apply certain style patterns to a given content image. The seminal work of Gatys~\etal~\cite{gatys2015neural} proposed to transfer the style of one image to another content image by optimizing the pixel values using both style and content losses,
inspiring many subsequent works in this field. To speed up the style transfer process, \cite{johnson2016perceptual} trained a feed-forward neural network for each style to transfer it to different contents. Going beyond single style transfer, \cite{huang2017arbitrary} introduced the idea of arbitrary style transfer and aimed to transfer arbitrary styles to any content in a single forward pass. Based on this formulation, \cite{li2019learning, park2019arbitrary, jing2020dynamic, liu2021adaattn, chen2021artistic, wu2022ccpl, sanakoyeu2018style, zhang2022domain} improved~\cite{huang2017arbitrary} on multiple aspects.

The above-mentioned methods can be classified as \textit{Image-guided Image Style Transfer (IIST)}. They rely on reference style images, which are not always accessible in real-world scenarios. 
For example, artists may conceive novel styles that can be easily described via texts but never exist in previous artworks. Such a dependence on reference style images limits the application of IIST methods \cite{kwon2022clipstyler}.

Recently, based on the large-scale image-text pretrained model CLIP~\cite{radford2021learning}, several methods proposed to edit images purely conditioned on text descriptions, achieving \textit{Text-guided Image Style Transfer (TIST)}~\cite{patashnik2021styleclip, kim2022diffusionclip, gal2021stylegan, kwon2022clipstyler}. Notably, by training a lightweight U-Net on a single content image using CLIP loss, CLIPStyler~\cite{kwon2022clipstyler} can synthesize stylized images from arbitrary content images and style text descriptions, setting a state-of-the-art (SOTA) performance on this task. However, although the styles of transferred images by CLIPStyler are generally consistent with the corresponding text descriptions, CLIPStyler often adds undesirable local patterns to the stylized images, distorting the original content severely, as shown in Figure~\ref{fig:cs_flaw}. This indicates CLIPStyler fails to disentangle the style and content information from both text and image. 

To address these issues, we propose a novel framework to better manipulate images based on reference style texts. Meanwhile, to offer more flexibility than IIST and TIST methods, our framework is designed to accept style guidance from multiple sources and modalities, enabling \textit{MultiModality-guided Image Style Transfer (MMIST)}.
The ability to exploit multimodal style references can be useful in many scenarios. For example, an artist may design new artistic styles by modifying styles of existing artworks; such modified styles can be easily defined by combining text descriptions and existing art images, yet are difficult to describe with text or image reference only.

To realize MMIST, we propose a novel cross-modal GAN inversion method which generates diverse style representations according to multi-modal style inputs (\textit{e.g.}, text and image). Such generated style representations allow us to generalize any IIST methods to tackle the problem of MMIST.
Specifically, we leverage a pretrained GAN model and invert style text descriptions and/or style images into GAN's latent space to get the corresponding style reference images. In this process, style-specific CLIP-based guidance is used to connect the domains of text and image. After obtaining the style reference images, we then feed them into 
an existing IIST approach which is adapted to take multiple style references as input.
To further enhance the quality of stylized images, we propose a novel multi-style boosting strategy which enriches the style patterns.
Similar to learned model parameters, the style representations can be reused at test time, allowing our method to  
stylize arbitrary contents in a single forward pass. 

We evaluate our framework on 44 style text descriptions and 61 content images, which result in 2,684 style-content combinations. Both qualitative and user study results clearly show the improvement of our framework over previous methods on TIST task. Furthermore, extensive experiments also confirm the effectiveness of our framework on MMIST task and cross-modal style interpolation. 

Our main contributions can be summarized as follows:
\begin{itemize}[itemsep=0.05cm]
    \item We introduce a more general problem than IIST and TIST, \ie, MultiModality-guided Image Style Transfer (MMIST),
    and solve it with a novel framework. The proposed framework can transfer styles from arbitrary number of reference images/texts to arbitrary content, a task which is not feasible for all existing methods to the best of our knowledge.

    \item We propose a novel cross-modal GAN inversion method to distill styles from different modalities. This inversion procedure also enables our method to interpolate between different styles arbitrarily.
    \item Extensive experiments and large-scale user studies (5,041 users) confirm the effectiveness of our model in terms of both qualitative results and user preference. 
\end{itemize}

\section{Related Work}

\subsection{Image-guided Image Style Transfer}

\begin{figure}[!t]
\centering
    \includegraphics[width=\linewidth]{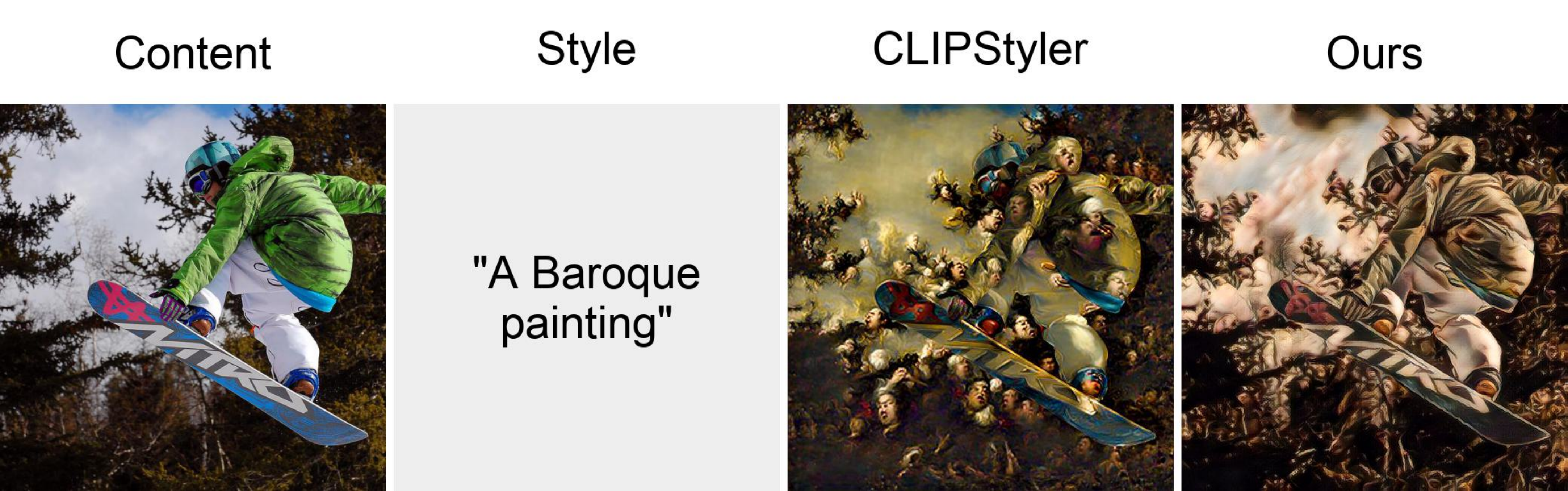}
    \caption{\textbf{Failure case of CLIPStyler.} We show one content and one style text description, together with the results from CLIPStyler and our method. CLIPStyler adds many small face-like patterns to the stylized images.}
\label{fig:cs_flaw}
\vspace{-0.3cm}
\end{figure}

Image-guided Image Style Transfer (IIST) has become a popular research topic since the seminal work of Gatys~\etal~\cite{gatys2015neural}. Using a deep feature-based style loss and content loss, Gatys~\etal~\cite{gatys2015neural} directly optimized pixel values to obtain decent style transfer results. To address the slow optimization problem in this method, \cite{johnson2016perceptual} and several subsequent works~\cite{ulyanov2016texture, ulyanov2017improved, wang2017multimodal} proposed to train feed-forward neural networks that can apply the pretrained style to arbitrary images in a single forward pass. Given its success, this idea was further developed by allowing one single trained model to store multiple styles. For example, ~\cite{chen2017stylebank} used multiple convolutional filter banks to explicitly represent multiple styles; \cite{dumoulin2016learned} proposed conditional instance normalization to achieve the same goal; and \cite{li2017diversified} introduced a selector structure to support incremental learning for new styles.

Recently, arbitrary IIST started to attract widespread attention due to its effectiveness, efficiency, and flexibility. \cite{huang2017arbitrary} proposed AdaIN to perform IST by adaptively aligning the mean and variance of content features with those of style features. 
With similar motivation, \cite{li2019learning, jing2020dynamic, park2019arbitrary, chen2021artistic} introduce new loss functions or novel mechanisms to improve the style transfer quality.
\cite{liu2021adaattn} 
adaptively performs attentive normalization on a per-point basis.
Besides quality, some works focus on other properties of IST methods, \eg, domain-awareness~\cite{hong2021domain} or brushstroke-level optimization~\cite{kotovenko2021rethinking}.\looseness=-1

\subsection{Text-guided Image Manipulation}

Text-guided image manipulation aims to manipulate the input image based on a text description while preserving text-irrelevant parts in the original image. \cite{dong2017semantic} employed a GAN-based encoder-decoder model to achieve this goal. \cite{nam2018text} further introduced a text-adaptive discriminator to ensure that only text-related regions are modified. By extending GAN-based text-to-image generators~\cite{zhang2017stackgan, zhang2018stackgan++, xu2018attngan}, 
Li~\etal \cite{li2020manigan} proposed ManiGAN to manipulate images in a multi-stage manner. \cite{xia2021tedigan} utilized GAN inversion, visual-linguistic similarity learning, and instance-level optimization to build a unified framework for multimodal image generation and manipulation with text.

Recently, the success of CLIP~\cite{radford2021learning} in connecting the domains of image and text has inspired a new direction to achieve text-guided image manipulation. By modifying the latent space of StyleGAN~\cite{karras2019style, karras2020analyzing, karras2021alias} using CLIP guidance, StyleCLIP~\cite{patashnik2021styleclip} can perform text-guided manipulation in three different ways. VQGAN-CLIP~\cite{crowson2022vqgan} achieved the same goal by using CLIP loss to optimize the latent space of VQGAN~\cite{esser2021taming}. StyleGAN-NADA\cite{gal2021stylegan} proposed a directional CLIP loss to optimize a GAN model instead of the latent space, resulting in more accurate manipulation effects. More recently, diffusion models~\cite{sohl2015deep, kingma2021variational, dhariwal2021diffusion, rombach2022high} have been combined with CLIP to obtain better performance~\cite{kim2022diffusionclip, avrahami2022blended}. 

Most of these works are intended for content or attribute editing. Although some of them~\cite{gal2021stylegan, kim2022diffusionclip} can be applied to style editing or transfer, the quality they can achieve is far from desirable. By contrast, CLIPStyler~\cite{kwon2022clipstyler} is specifically designed to solve the task of TIST, outperforming all previous image manipulation methods on this task. However, CLIPStyler still suffers from certain drawbacks, as illustrated in Figure~\ref{fig:cs_flaw}. Similar to CLIPStyler, \cite{liu2022name} proposed to transfer an image's style conditioned on an artist's name.  
In this paper, we generalize the tasks of TIST and IIST to MMIST, and solve it under a unified framework.

\subsection{GAN Inversion}

Traditional GAN inversion aims to invert a given image back into the latent space of a pretrained GAN generator.
It emphasizes the accuracy and fidelity of the reconstructed image.
GAN inversion can be applied to a wide range of downstream tasks, including image manipulation~\cite{patashnik2021styleclip, gal2021stylegan, xia2021tedigan}, image interpolation~\cite{abdal2019image2stylegan}, image generation~\cite{xia2021tedigan}, etc. Learning-based~\cite{zhu2016generative}, optimization-based~\cite{abdal2019image2stylegan}, or hybrid~\cite{alaluf2022hyperstyle} methods have been developed to invert a GAN. All of them use a reconstruction loss such as $L_2$ loss or LPIPS loss~\cite{zhang2018unreasonable}.  \looseness=-1

In this paper, we propose cross-modal GAN inversion. Different from traditional methods that pursue a perfect reconstruction of the whole input image, our cross-modal GAN inversion only reconstructs partial information of the input, \ie, style,  which is defined by inputs of multiple modalities such as text and image.

\section{Method Overview}

\begin{figure}[!t]
\centering
    \includegraphics[width=\linewidth]{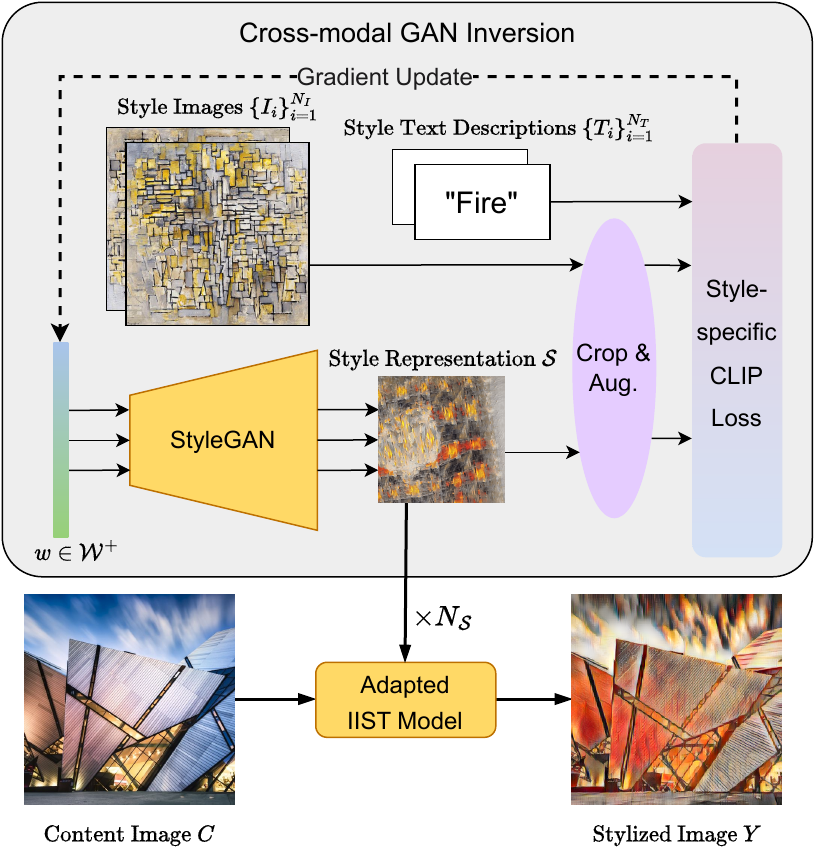}
    \caption{\textbf{The overview of our method.} Taking styles from multiple modalities as input, our method generates style representations using cross-modal GAN inversion.
    With the adapted IIST model, our method can apply the cached styles to any unseen content image in a single forward pass.
    }
\label{fig:overview}
\vspace{-0.3cm}
\end{figure}

Given a set of style images $\isi$ and a set of style text descriptions $\ist$, our framework applies these specified styles to a set of content images $\{C_i\}_{i=1}^{N_C}$ and synthesizes a corresponding set of stylized images $\{Y_i\}_{i=1}^{N_C}$. Different from previous image editing methods that directly optimize the stylized image~\cite{kwon2022clipstyler}, the latent of the content image~\cite{patashnik2021styleclip, crowson2022vqgan}, or parameters of a generative model~\cite{gal2021stylegan, kim2022diffusionclip}, we train a model for certain styles in a content-agnostic manner.

Our key insight is that MMIST can be achieved with the aid of style representations that comply with the input style text descriptions and image patterns. More specifically, we can generalize IIST methods to leverage such style representations and thereby create stylized images guided by multiple modalities.
To this end, 
we propose a novel cross-modal GAN inversion method to map all input multimodal style references into the $\mathcal{W}^+$ space~\cite{abdal2019image2stylegan} of a pretrained StyleGAN3~\cite{karras2021alias} generator $\G$, and thereby generate intermediate style representations $\sr$ in the image space. As is shown in Figure~\ref{fig:overview},
the style representations are images consisting of style \textit{patterns} without meaningful contents. 
The cross-modal GAN inversion ensures that $\sr$ summarizes and combines the style information from all input styles $\isi$ and $\ist$. 
To leverage $\sr$, we adapt an existing IIST method to make it compatible with multiple style inputs.
Denote the adapted IIST method by $\adaattn$. We use $\adaattn$ to stylize the content images $\{C_i\}_{i=1}^{N_C}$ with intermediate style representations $\sr$, producing stylized outputs $\{Y_i\}_{i=1}^{N_C}$.

Separating style representation generation from stylized image synthesis is the key to the success of our framework. When performing TIST task, previous methods~\cite{crowson2022vqgan, kwon2022clipstyler, kim2022diffusionclip} always deal with style alignment and content preservation simultaneously, resulting in distorted content or irrelevant artifacts that appear in the results. In contrast, by leveraging the strong style-content disentangling ability of IIST approaches, our method can put the entire focus on style representation generation for creating high-quality stylized images. Besides, with generated intermediate style representations, only a single forward pass is needed for our method to apply a learned style to any unseen content image. 

\section{Cross-modal GAN Inversion}

To generate style representations from modalities other than image or text only, we propose cross-modal GAN inversion. In Table~\ref{tab:comp}, we compare it with traditional GAN inversion. The goal of traditional GAN inversion is to faithfully reconstruct the original input image. Naturally, this only works for the image modality, and only one image at a time. Since it is targeting pixel-wise reconstruction, all information from the input image is supposed to be stored in the latent space of the GAN generator. However, the goal of cross-modal GAN inversion is completely different. 
It aims to combine different styles together to generate intermediate style representations. Therefore, the inversion should be able to accept multiple inputs from different references and modalities. Besides, only the style components of inputs are required to be inverted as their content parts are irrelevant to the downstream task.

\subsection{Style-specific CLIP Loss}

We employ CLIP~\cite{radford2021learning} to connect image with other modalities, as well as to extract style components. However, naively applying CLIP cosine similarity loss does not result in accurate style representation, since the content components are also entangled in the CLIP embedding space. 

Following~\cite{kwon2022clipstyler} and~\cite{gal2021stylegan}, we employ patch-wise CLIP loss to address this problem. 
Formally, denote the pretrained CLIP image encoder by $\ei$ and text encoder by $\et$. For each style text description $\ti$, we use the text-image patch-wise directional CLIP loss proposed by~\cite{kwon2022clipstyler}, \ie,
\begin{align}
\begin{split}
    \label{eq:tiloss}
    &\mathcal{S} = \G(w), \\
    &\{\mathcal{S}^j\}_{j=1}^{N_\text{crop}} = \aug(\crop(\mathcal{S})), \\
    &\Delta \mathcal{S}^j = \ei(\mathcal{S}^j) - \ei(I_\text{src}), \\
    &\Delta T = \et(\ti) - \et(T_\text{src}), \\
    &L_{\ti} = \frac{1}{N_\text{crop}} \sum_{j=1}^{N_\text{crop}} \left( 1 - \frac{\Delta \mathcal{S}^j \cdot \Delta T}{\|\Delta \mathcal{S}^j\|\|\Delta T\|} \right),
\end{split}
\end{align}
where $w \in \mathcal{W}^+$ is a vector in StyleGAN3 latent space that we optimize, $\mathcal{S}$ is the style representation generated by $\G$ from $w$. $\aug(\cdot)$ is the augmentation function, $\crop(\cdot)$ is the patch crop function,  and $N_\text{crop}$ is the number of cropped patches.
$T_\text{src}$ and $I_\text{src}$ are the source text and source image used to compute CLIP embedding directions, respectively. For simplicity, $T_\text{src}$ is set to be ``a photo'' following~\cite{kwon2022clipstyler}, whereas $I_\text{src}$ is an arbitrary photo-realistic image.

\begin{table}[t]
    \centering
    \caption{\textbf{Comparison between traditional GAN inversion and cross-modal GAN inversion.} ``Ref." means reference.}
    \resizebox{0.475\textwidth}{!}{
    \begin{tabular}{@{}lccc@{}}
        \toprule
        Method & Ref. Modality & Number of Ref. & Inversion Target \\
        \midrule
        Traditional & Image Only & Single & Original Image \\
        Cross-modal & Multiple & Multiple & Style \\
        \bottomrule
    \end{tabular}
    }
    \label{tab:comp}
    \vspace{-0.7cm}
\end{table}

Eq.~\ref{eq:tiloss} effectively measures the style similarity between the input text $\ti$ and the generated style $\mathcal{S}$. However, we want our model to handle style inputs from the image modality as well. To this end, we propose an image-image patch-wise directional CLIP loss as below:
\begin{align}
\begin{split}
    \label{eq:iiloss}
    &\{\ii^k\}_{k=1}^{N_\text{crop}} = \aug(\crop(\ii)), \\
    &\Delta \ii^k = \ei(\ii^k) - \ei(I_\text{src}), \\
    &L_{\ii} = \frac{1}{N_\text{crop}^2} \sum_{j=1}^{N_\text{crop}}\sum_{k=1}^{N_\text{crop}} \left( 1 - \frac{\Delta \mathcal{S}^j \cdot \Delta \ii^k}{\|\Delta \mathcal{S}^j\|\|\Delta \ii^k\|} \right),
\end{split}
\end{align}
where $\ii$ is the input style image, and $\Delta \mathcal{S}^j$ is calculated using Eq.~\ref{eq:tiloss}. 
Specifically, for every image $\ii$ or $\si$ involved in the CLIP embedding computation, we first randomly crop a large number of patches and then augment them. After computing the loss for each patch, we average them together to obtain the final loss value. 
By computing the averaged cosine similarity between each direction pair, Eq.~\ref{eq:iiloss} accurately estimates the style similarity between the input image $\ii$ and the generated style $\mathcal{S}$.

\begin{algorithm}[t!]
    \SetAlgoLined
    \KwData{A set of style images $\isi$, a set of style text descriptions $\ist$, and corresponding style weights $\{\alpha_i^I\}_{i=1}^{N_I}$, $\{\alpha_i^T\}_{i=1}^{N_T}$. }
    \KwResult{The generated style representation $\mathcal{S}$ and its corresponding latent $w^* \in \mathcal{W}^+$.}
    
    Randomly initialize $w$;
    
    \Repeat{$L_\text{sty}$ is converged}{
        $\mathcal{S} \gets \G(w)$\; 
        Run $\aug$ and $\crop$ on $\mathcal{S}$ and $\isi$ to obtain $\{\mathcal{S}^j\}_{j=1}^{N_\text{crop}}$, $\{I_i^k\}_{i, k=1, 1}^{N_i, N_\text{crop}}$\;
        Calculate $L_\text{sty}$ using Eq.~\ref{eq:tiloss}, Eq.~\ref{eq:iiloss}, and Eq.~\ref{eq:opt}\;
        Adam update for $w$ with $\nabla_{w} L_\text{sty} $\;
        }
    \Return{$(\mathcal{S}, w^*)$}\;
        
    \caption{Cross-modal GAN Inversion}
    \label{alg:cmgi}
\end{algorithm}

In the general case where multiple style images $\isi$ and style text descriptions $\ist$ are given, we calculate the style-specific CLIP loss $L_\text{sty}$, and solve the following optimization problem:
\begin{align}
\label{eq:opt}
    w^* = \argmin_{w \in \mathcal{W}^+} L_\text{sty} =  \argmin_{w \in \mathcal{W}^+} \sum_{i=1}^{N_I} \alpha_i^I L_{\ii} + \sum_{i=1}^{N_T} \alpha_i^T L_{\ti},
\end{align}
where $\{\alpha_i^I\}_{i=1}^{N_I}$ and $\{\alpha_i^T\}_{i=1}^{N_T}$ are the style weights.

\begin{algorithm}[t!]
    \SetAlgoLined
    \KwData{Input styles $\isi$, $\ist$, style weights $\{\alpha_i^I\}_{i=1}^{N_I}$, $\{\alpha_i^T\}_{i=1}^{N_T}$, and a set of content images $\{C_i\}_{i=1}^{N_C}$. }
    \KwResult{A set of stylized images $\{Y_i\}_{i=1}^{N_C}$.}
    \If{The aggregated feature $F$ is not cached}
    {
        Run Algorithm~\ref{alg:cmgi} $N_\mathcal{S}$ times to obtain $\sr$\;
        $\{F_i\}_{i=1}^{N_C} \gets \adaattn_f(\sr)$\;
         $F \gets \concat(\{F_i\}_{i=1}^{N_C})$\;
    }
    $\{Y_i\}_{i=1}^{N_C} \gets \adaattn_t(\{C_i\}_{i=1}^{N_C}, F)$\;
    \Return{$\{Y_i\}_{i=1}^{N_C}$}\;
    \caption{Multimodality-guided Style Transfer}
    \label{alg:st}
\end{algorithm}

\subsection{Inversion Algorithm}
\label{sec:invalg}
With the style-specific CLIP guidance, it is straightforward to run our cross-modal GAN inversion algorithm. As shown in Algorithm~\ref{alg:cmgi}, after initializing $w$, we repetitively calculate $L_\text{sty}$ and use Adam optimizer~\cite{kingma2014adam} to update $w$, until $L_\text{sty}$ has converged.

Note that in this algorithm, both $\isi$ and $\ist$ are optional. If only $\ist$ is given, it degenerates to text-guided style generation, 
converting our framework to a method for TIST.
Similarly, if only $\isi$ is given,
it degenerates into mixing multiple style images, \ie, 
generating a mixture of styles represented by multiple input images. 
 When both $\isi$ and $\ist$ are provided, cross-modal style interpolation, \eg, interpolating a style between a given text and a given image, can be naturally achieved by adjusting the style weights $\{\alpha_i^I\}_{i=1}^{N_I}$ and $\{\alpha_i^T\}_{i=1}^{N_T}$. 

\begin{figure*}[!t]
\centering
    \includegraphics[width=\linewidth]{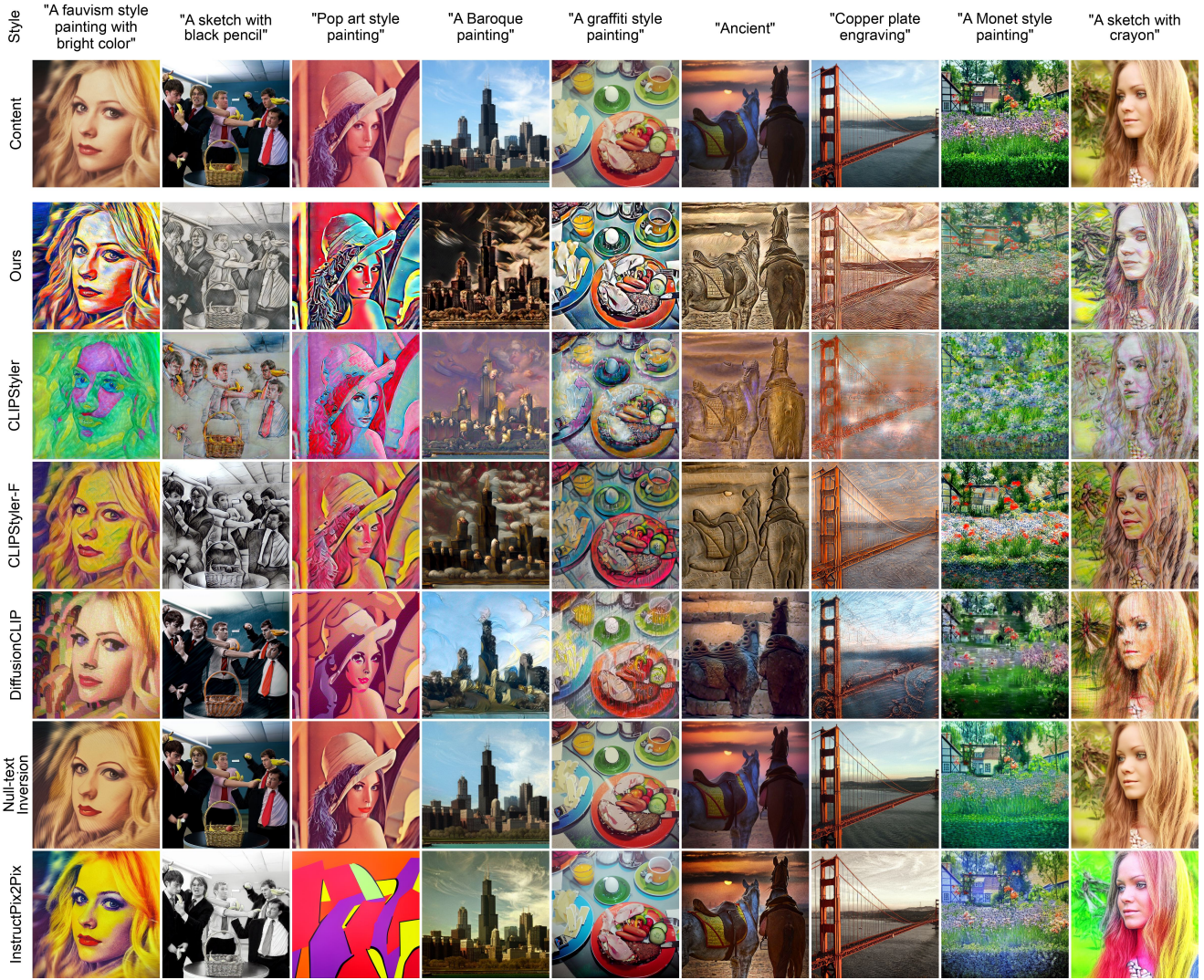}
    \caption{\textbf{Comparison with other TIST methods.} Our method delivers more accurate styles than others. Meanwhile, as opposed to other methods which often distort the content or add unreasonable patterns, 
    content information is greatly preserved in results of our method.
    }
    \vspace{-0.5cm}
\label{fig:text_comp}
\end{figure*}

\section{Multimodality-guided Image Style Transfer}

\subsection{Multi-style Boosting}

Due to the internal randomness of GAN inversion, one single intermediate style representation may not cover all style patterns specified by the input references, impairing the style transfer quality of final results. To address this problem, we propose a multi-style boosting algorithm. We aim to enrich the intermediate style representations, while keeping them compatible with the adapted IIST model $\adaattn$. Specifically, for each set of style inputs, we run cross-modal GAN inversion multiple times, resulting in a set of style representations $\sr$. 
Then we feed them into $\adaattn$ separately, and aggregate the outputs together to exploit $\sr$. The aggregation strategy depends on the specific implementation of $\adaattn$. We describe one instance of $\adaattn$ and the aggregation strategy in the  supplementary material.

\subsection{Style Transfer Algorithm}
\label{sec:algo}

We detail our style transfer method in Algorithm~\ref{alg:st}. For brevity, 
we assume
the adapted IIST model $\adaattn$
can be decomposed into a feature extraction network $\adaattn_f$ and a style transfer module $\adaattn_t$, \ie, $\adaattn(C, \mathcal{S}) = \adaattn_t (C, \adaattn_f(\mathcal{S}))$. 
Similar to Algorithm~\ref{alg:cmgi}, Algorithm~\ref{alg:st} can be used for TIST by only providing $\ist$, or MMIST by providing both $\isi$ and $\ist$.
In particular, if only one $T_0$ is given, our method degenerates to the IIST approach it generalizes since cross-modal GAN inversion is not necessary in this case.
The aggregated style feature $F$ from each unique set of input styles can be cached for later use once it is produced.
With cached $F$, when new content images arrive with the same input styles, only the style transfer part $\adaattn_t$ needs to be executed. Since $\adaattn_t$ is a feed-forward network, our method runs significantly faster than previous methods~\cite{kwon2022clipstyler, kim2022diffusionclip, crowson2022vqgan} that require ad-hoc optimization for each pair of style and content. 

\vspace{-0.2cm}
\section{Experiments}
\label{exp}
We consider three tasks to evaluate our method: (1) TIST; (2) MMIST with one style image and one style text description; and 
(3) MMIST with style interpolated from multiple references in different modalities.
Note that IIST is trivial and unnecessary to consider because our method degenerates to the IIST approach it includes, as mentioned in Section~\ref{sec:algo}. We employ AdaAttN~\cite{liu2021adaattn} as our adapted IIST method in all our experiments. Please see the supplementary material for more implementation details.

\subsection{Comparison with TIST Methods}

\noindent\textbf{Qualitative comparison.} We conducted a qualitative comparison between our method and several TIST methods. In this task, style transfer is conditioned on a single text description. Therefore, we simply set 
$N_T = 1$ and $N_I = 0$ 
in our method. We consider CLIPStyler~\cite{kwon2022clipstyler}, CLIPStyler-F~\cite{kwon2022clipstyler}, DiffusionCLIP~\cite{kim2022diffusionclip}, Null-text Inversion~\cite{mokady2023null}, and InstructPix2Pix~\cite{brooks2023instructpix2pix} as our baselines. These methods either exclusively focus on this task~\cite{kwon2022clipstyler} and achieve state-of-the-art performance, or treat TIST as one of their practical applications. It is worth noting that CLIPStyler and CLIPStyler-F are proposed in the same paper, and the latter is an extension of the former and achieves TIST in a single forward pass for learned style text descriptions. 
By caching style representations, the speed of our method is similar to CLIPStyler-F, which is significantly faster than CLIPStyler and DiffusionCLIP.

\begin{table}[t]
    \centering
    \caption{\textbf{Quantitative user study results on four TIST methods.} For each method, we report the number of positive responses received from raters, as well as its percentage (over 161,040 responses in total). Our method outperforms baselines.}
    \resizebox{0.475\textwidth}{!}{
    \begin{tabular}{@{}lrrr@{}}
        \toprule
        Method & Style (\%) $\uparrow$ & Content (\%) $\uparrow$ & Overall (\%) $\uparrow$ \\
        \midrule
        CLIPStyler & 10631 (39.6) & 9686 (36.1) & 8966 (33.4) \\
        CLIPStyler-F & 14626 (54.5) & 13453 (50.1) & 11776 (43.9) \\
        DiffusionCLIP & 9768 (36.4) & 7929 (29.5) & 7395 (27.6) \\
        Null-text Inversion & 7375 (27.5)  & 16876 (62.8) & 7166 (26.7) \\
        InstructPix2Pix & 10580 (39.4) & 17453 (65.0) & 8268 (30.8) \\
        Ours & \textbf{17151} (\textbf{63.9}) & \textbf{18061} (\textbf{67.3}) & \textbf{15125} (\textbf{56.4}) \\
        \bottomrule
    \end{tabular}
    }
    \label{tab:comp_user_study}
    \vspace{-0.6cm}
\end{table}

Figure~\ref{fig:text_comp} shows comparison results on 9 content-style pairs. We list the style text descriptions and the content image in the first and the second rows, respectively. Style transfer results of different methods are listed in the remaining rows. 
As is shown, CLIPStyler often breaks or distorts the original contents (1st and 7th columns). For results generated by DiffusionCLIP, their styles do not match the corresponding text descriptions well (1st, 2nd, and 7th columns). In addition, these methods tend to add undesirable local patterns to stylized images (4th and 9th columns). 
The absence of an original image caption likely explains why Null-text Inversion's outcomes closely mirror the content images, displaying minimal style variations. InstructPix2Pix sometimes compromises the content (3rd column) or introduces inaccurate styles (2nd and 7th columns).
In contrast, our method delivers more accurate styles while greatly preserving the content information.

\noindent\textbf{Quantitative user study.} 
We conduct a large-scale quantitative user study to better understand the performance of our method. We still use CLIPStyler~\cite{kwon2022clipstyler}, CLIPStyler-F~\cite{kwon2022clipstyler}, DiffusionCLIP~\cite{kim2022diffusionclip}, Null-text Inversion~\cite{mokady2023null}, and InstructPix2Pix~\cite{brooks2023instructpix2pix} as our baselines. We apply 44 distinctive text-described styles to 61 different content images, giving 2,684 stylized images. For each of them, we ask 10 different raters to evaluate it from three aspects: style consistency, content preservation, and overall quality. For each aspect, raters are asked if the stylized image \textit{respects the aspect well} (positive) or \textit{not} (negative). 
In total we obtain 161,040 responses where each method receives 26,840 responses and each stylized image receives 10 responses. 
The total number of raters involved is 5,041.
We report the number and the percentage of positive responses. Table~\ref{tab:comp_user_study} shows that our method outperforms all baselines in every aspect. 
Interestingly, CLIPStyler-F received more positive responses than CLIPStyler, although the former is designed to be a fast extension in~\cite{kwon2022clipstyler}. 
This user study result is consistent with the qualitative results shown in Figure~\ref{fig:text_comp}.

\noindent\textbf{Running speed.} Our pre-computed style representations enable \textit{fast stylization} at \textit{test time}. 
We compare speed of our method with other TIST methods in Table~\ref{tab:runtime}. All methods are under their fastest setting and run on an RTX A6000.

\begin{table}[t]
    \centering
    \caption{\textbf{Speed comparison with other TIST methods.} Our method is the fastest. ($\star$) Null-text Inversion does not need per-style training but has a per content inverison time of 80.7s.}
    \resizebox{0.45\textwidth}{!}{
    \begin{tabular}{@{}lcc@{}}
        \toprule
        Method & Per-style Training Time (s) $\downarrow$ &  Stylization FPS $\uparrow$ \\
        \midrule
        CLIPStyler & 0 & 0.03  \\
        CLIPStyler-F & 44.1 & 4.40 \\
        DiffusionCLIP & 1885 & 0.05 \\
        Null-text Inversion & $\star$ & 0.07 \\
        InstructPix2Pix & 0 & 0.08 \\
        Ours & 12.6 & \textbf{6.00}  \\
        \bottomrule
    \end{tabular}
    }
    \label{tab:runtime}
    \vspace{-0.3cm}
\end{table}

\vspace{-0.1cm}
\subsection{MMIST and Cross-modal Style Interpolation}

\begin{figure*}[!t]
\centering
    \includegraphics[width=0.968\linewidth]{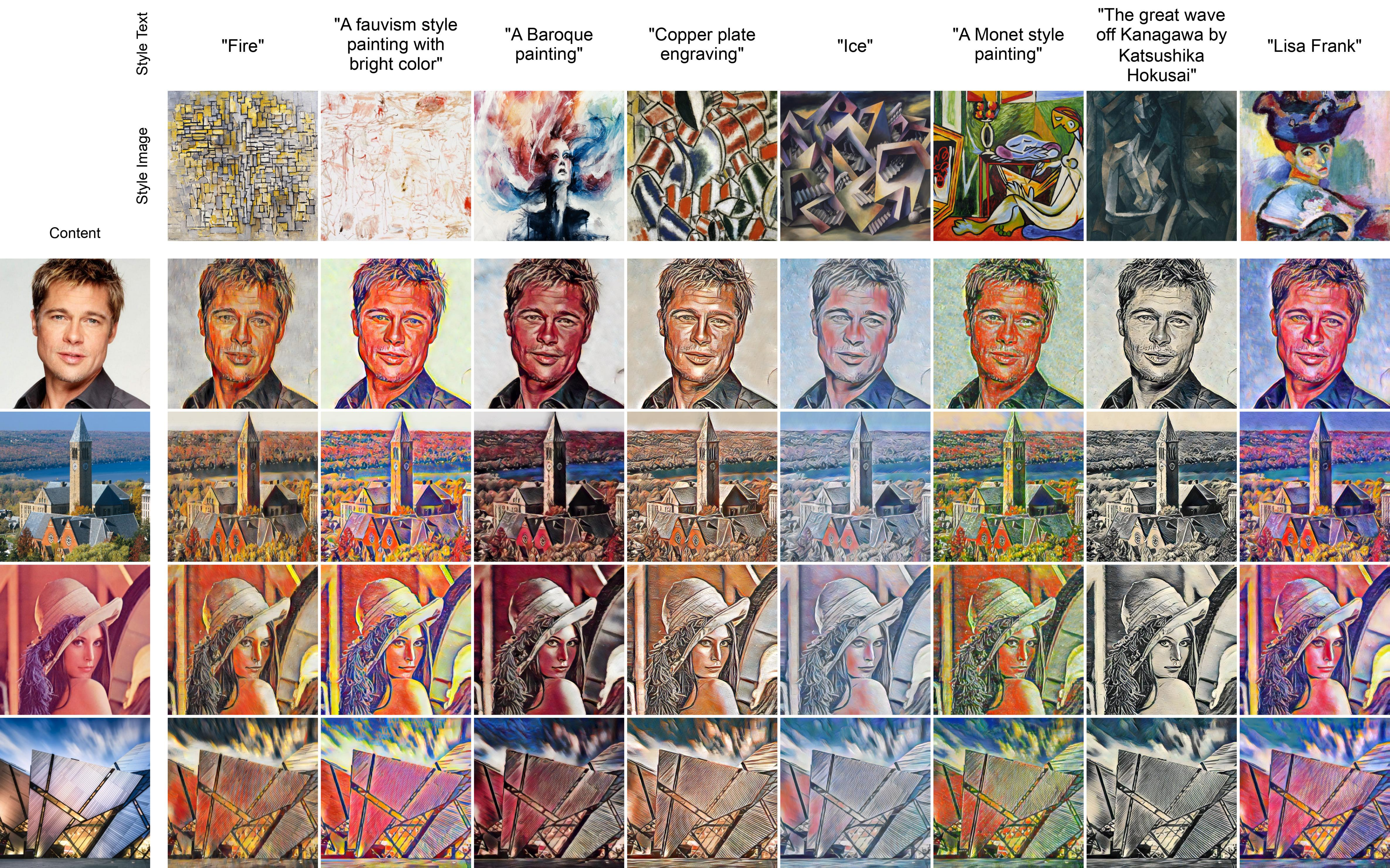}
    \caption{\textbf{MMIST results.}
    The first and second rows show style text descriptions and style images, respectively. The first column shows  content images. 
    Our method successfully mixes multimodal styles and applies them to various content images.
    }
    \vspace{-0.5cm}
\label{fig:multi_results}
\end{figure*}

\noindent\textbf{Qualitative results of MMIST with one style image and one style text
description.} 
To the best of our knowledge, MMIST is infeasible for all the existing methods. However, under our unified framework, it can be easily performed by providing style references from more than one modality.
We first consider the case where input style is specified through one style image and one style text description.
Figure~\ref{fig:multi_results} shows MMIST results of our method on 8 different text-image styles and 4 content images. 
Our method can successfully summarize styles from a pair of style image and text description, and apply it to various contents. Moreover, contents are consistently preserved in all stylized images.

\begin{figure}[t]
\centering
    \includegraphics[width=\linewidth]{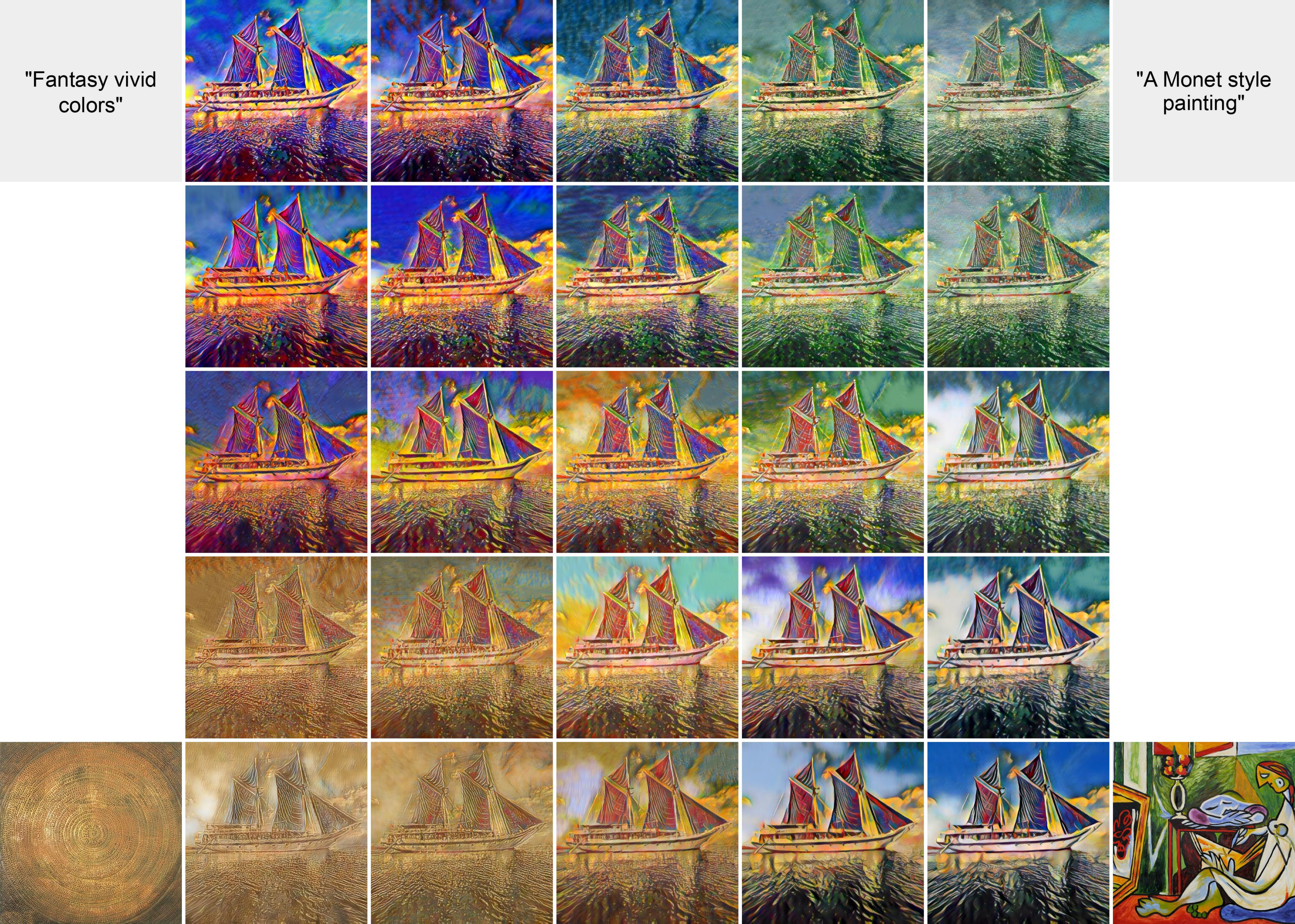}
    \caption{\textbf{MMIST and Style interpolation with four image and text styles.} Please zoom in to see the details. }
\label{fig:four_results}
\vspace{-0.5cm}
\end{figure}

\noindent\textbf{Qualitative results of MMIST with style interpolated from multiple references in different modalities.} 
As mentioned in Section~\ref{sec:algo}, our method is able to handle arbitrary number of style inputs from different modalities. 
To finer control the mixture degree of two or more given styles, we interpolate the styles with various ratios. This is done by setting different $\{\alpha_i^I\}_{i=1}^{N_I}$ and $\{\alpha_i^T\}_{i=1}^{N_T}$, while keeping their summation fixed. Note that under our unified framework, style interpolation between any number or kinds of modalities can be achieved in the same way.
Figures~\ref{fig:teaser} and \ref{fig:four_results} demonstrate the style interpolation results defined by different mixtures of style images and text descriptions. Our method produces decent and reasonable stylized images.
More results can be found in the supplementary material.

\subsection{Ablation Study}

\begin{table}[t]
    \centering
    \caption{\textbf{Ablation study on different design choices.} The performance is evaluated through the user study.}
    \resizebox{0.475\textwidth}{!}{
    \begin{tabular}{@{}lc|lc@{}}
        \toprule
        Setting & Preference \% $\uparrow$ & Setting & Preference \% $\uparrow$\\
        \midrule
        CropSize128 & 39.5 & PatchLoss500 & 49.9 \\
        NoCrop & 41.9 & PatchLoss2500 & 48.2 \\
        NoAug & 47.9 & NoBoosting & 34.5  \\          
        \bottomrule
    \end{tabular}
    }
    \label{tab:ab_user_study}
    \vspace{-0.2cm}
\end{table}

We investigate a few design choices of our method and quantitatively measure their effects in practice.
To this end, we consider the TIST task. In our experiment, we randomly pair the style text descriptions and content images, and obtain 1,012 pairs for stylized image generation. For each specific pair, 10 raters are asked to pick their preferred stylized image between the one generated by a certain design choice and the one generated by our full method. We report the user preference percentage for all design choices in Table~\ref{tab:ab_user_study}. In the table, CropSize128 means we crop $128 \times 128$ patches when calculating style-specific CLIP loss. NoCrop means we do not crop or augment the image, and directly apply the loss. NoAug means we do not apply augmentation to the cropped patches. PatchLoss500 and PatchLoss2500 mean we set $\alpha_0^T = 500$ and $\alpha_0^T = 2500$, respectively. NoBoosting means we do not use multi-style boosting when applying the style to contents. We observe that proper crop size and multi-style boosting are critical to the good performance of our method, whereas the effects of augmentation and style weight are relatively minor to our method. Changing the style weight from 500 to 2500 almost does not affect the performance, indicating that our method is quite robust to this hyperparameter.  Qualitative ablation study results are available in the supplementary material. \looseness=-1

\section{Conclusions and Future Work}
In this paper, we present a unified style transfer framework to transfer styles defined by multiple modalities. The proposed cross-modal GAN inversion enables our framework to combine different styles and faithfully transfer them to arbitrary images.
Extensive experiments demonstrate that our method achieves SOTA performance on TIST. In addition, the proposed method handles the new MMIST problem and cross-modal style interpolation task effectively.

While our work only considers style information from image and text, there is no theoretical restriction on our method to obtain styles from other modalities, \eg, audio. We leave the exploration of this direction as future work.

\clearpage

{\small
\bibliographystyle{ieee_fullname}
\bibliography{egbib}
}

\clearpage
{\noindent\huge \textbf{Supplementary Material}}

\section{Implementation Details}

For all of our cross-modal GAN inversion experiments, we utilize a pretrained StyleGAN3-T model~\cite{karras2021alias} that was trained on the WikiArt dataset\footnote{\url{https://www.wikiart.org/}}.
We use this\footnote{\url{https://github.com/Huage001/AdaAttN}} implementation in all our experiments.
It is worth noting that the performance of this StyleGAN3-T model may be restricted by its pretraining dataset, which only involves the WikiArt dataset. Nevertheless, despite using this domain-limited StyleGAN3-T, our approach still remains competitive, as evidenced by our qualitative findings and user study. By employing more powerful generators, our approach can achieve even better performance.

For cross-modal GAN inversion, we use Adam optimizer~\cite{kingma2014adam} with a learning rate of $0.2$. We set the summation of all style weights $\{\alpha_i^I\}_{i=1}^{N_I}$ and $\{\alpha_i^T\}_{i=1}^{N_T}$ to be $1000$.

To make fair comparisons with previous works, we set the spatial resolution to $512 \times 512$ for all image data in our framework. We use a patch size of $256$ in all patch-wise CLIP losses. 
To compute the proposed style-specific CLIP loss, we use the CLIP ViT-B/32~\cite{radford2021learning} model.
Following~\cite{patashnik2021styleclip, gal2021stylegan, kim2022diffusionclip, kwon2022clipstyler}, we apply prompt augmentation~\cite{radford2021learning} to all text descriptions by default.
When computing this loss, we resize all inputs to $224 \times 224$ to make them compatible with the image encoder of CLIP. Following \cite{kwon2022clipstyler}, we apply random perspective augmentation with a distortion scale of $0.5$ to all image data used in our main results. In other words, the $\aug(\cdot)$ function defined in our main paper is implemented as RandomPerspective(fill=0, p=1, distortion\_scale=0.5) using $\mathrm{torchvision.transforms}$. It takes 20 iterations to run our cross-modal GAN inversion. Our complete code will be made available.

\section{Style Text Descriptions and Content Images}
We use $11$ content images and $20$ style images released by~\cite{huang2017arbitrary}. We also use $50$ square-shaped images randomly sampled from COCO test set~\cite{lin2014microsoft} as a supplement to our content set.
In addition, we manually collect $44$ style text descriptions, including those used by~\cite{kwon2022clipstyler}.
We list all style text descriptions in the attached file, \textit{style\_text.txt}. And we put all content images in the \textit{content} folder.

\section{User Study Design}

In our user studies, we ask professional annotators from Scale AI\footnote{\url{https://scale.com/}} to evaluate all our results.

In the main user study (\textit{i.e.}, Table 2 in the main paper),  we apply 44 distinctive text-described styles to 61 different content images, giving 2,684 stylized images. For each of them, we ask $10$ different annotators to evaluate it from three aspects: style consistency, content preservation, and overall quality. For each aspect, annotators are asked if the stylized image \textit{respects the aspect well} (positive) or \textit{not} (negative). 
In total we obtain 26,840 responses, where each stylized image received 10 responses. 

In our user study for ablation study, we randomly pair the style text descriptions and content images. Specifically, we use all $44$ style text descriptions, and pair each of them with $23$ randomly sampled content images for style transfer, giving 1,012 stylized images.

In Table~\ref{tab:user_study_design}, we list the number of annotators involved in each evaluation task. As is shown, our user study is based on a sufficiently large number of annotators.

In Figures~\ref{fig:supp_anno_style}, \ref{fig:supp_anno_content}, \ref{fig:supp_anno_overall}, and \ref{fig:supp_anno_ab}, we show the annotation user interfaces for evaluating style consistency, content preservation, overall quality, and ablation methods, respectively.
In addition, we further show several annotation examples in Figures~\ref{fig:supp_anno_eg_style}, \ref{fig:supp_anno_eg_content}, and \ref{fig:supp_anno_eg_overall}. The number of positive responses received for these images is listed at the bottom. We observe that the evaluation results from annotators are reasonable and consistent with the quality of stylized images.

\begin{figure*}[h]
    \centering
    \includegraphics[width=\linewidth]{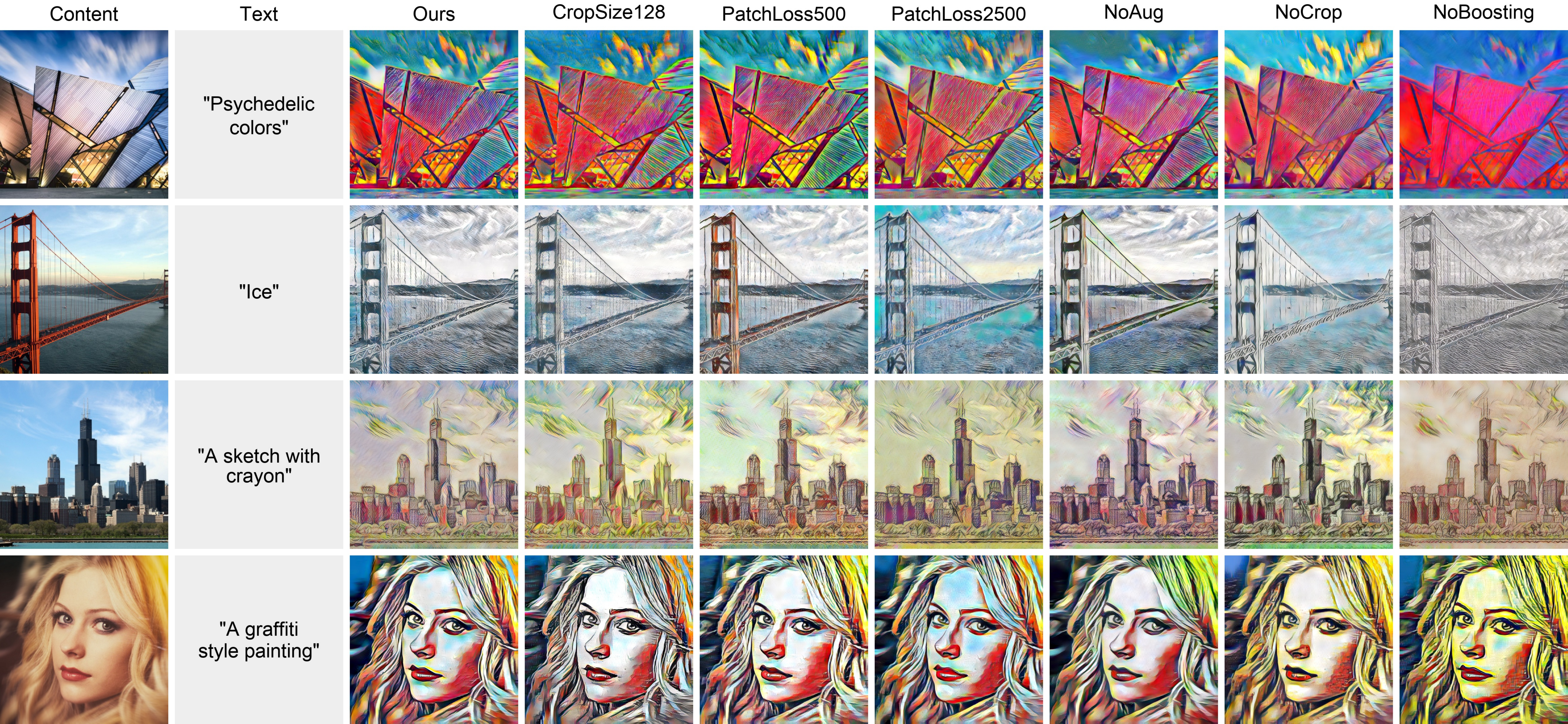}
    \caption{\textbf{Qualitative ablation study on different design choices.} We compare  our final method with all design choices used in Table~4 of the main paper. Zoom in for a better view.}
    \label{fig:supp_ab_qual}
\end{figure*}

\begin{figure}[!h]
\centering
    \includegraphics[width=\linewidth]{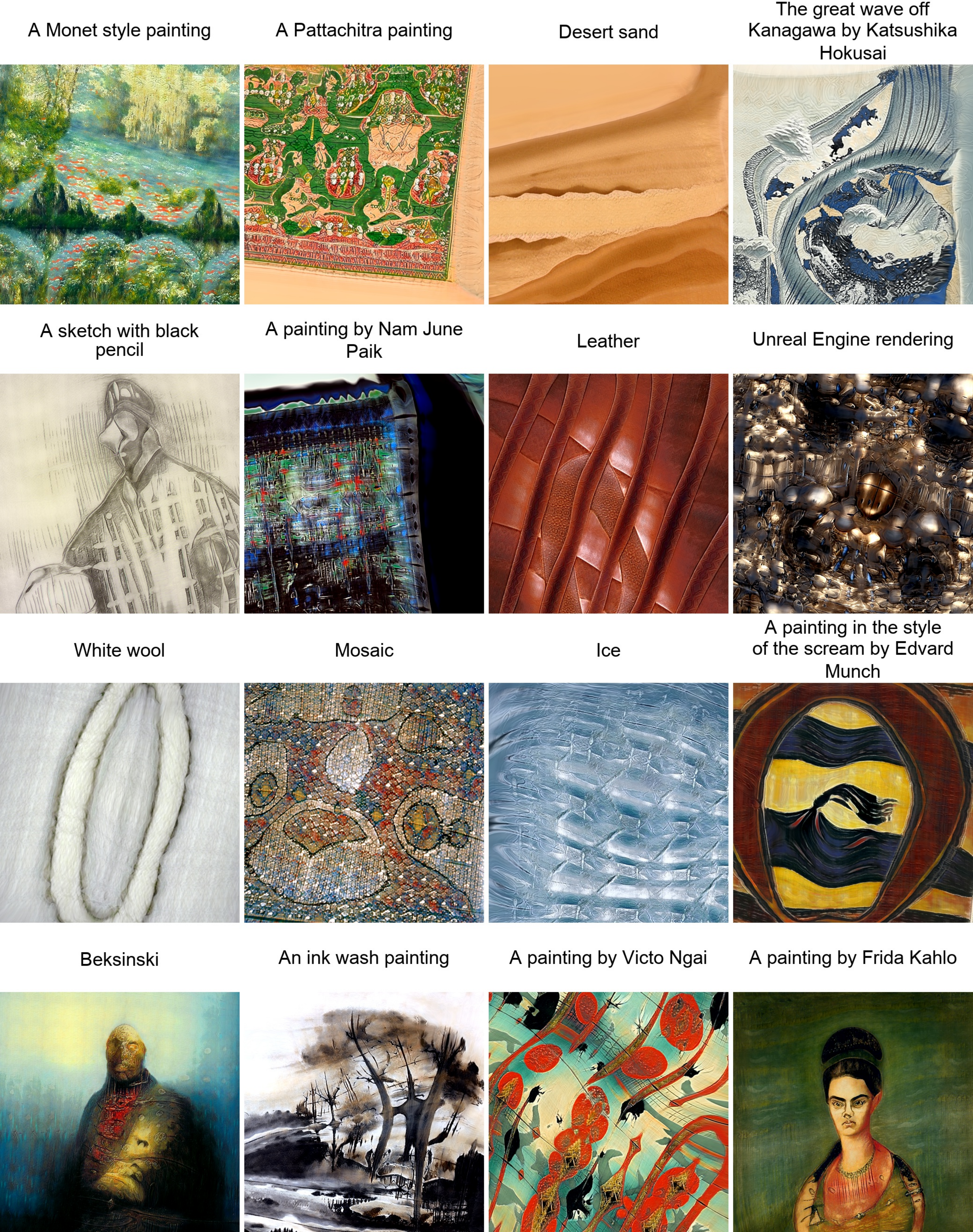}
    \caption{\textbf{Inverted style representation examples.} The corresponding style text description is displayed above each style representation.
    }
\label{fig:inv_examples}
\end{figure}

\begin{figure*}[!h]
\centering
    \includegraphics[width=\linewidth]{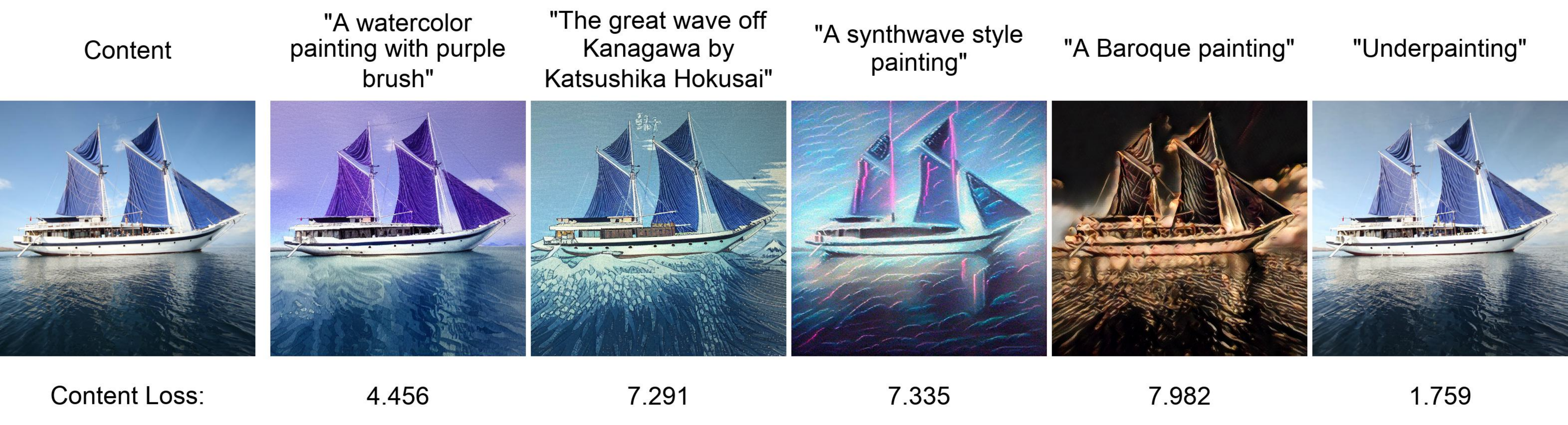}
    \caption{\textbf{Problem of the content loss.} This figure shows randomly picked stylized images and their content loss values. Note that they are obtained from \textbf{different randomly selected style transfer methods}. The method names are intentionally hidden to ensure unbiased perception. The fisrt image is the original content image and the remaining ones are the stylized images. Style text descriptions are shown above the images. Content loss values are shown below the images. The highly stylized images (2nd and 4th) appear to incur a higher content loss, even though they largely preserve the original content. In contrast, the 5th stylized image, which deviates minimally from the content image, achieves the best content loss.
    }
\label{fig:content_loss_1}
\end{figure*}

\begin{figure*}[!h]
\centering
    \includegraphics[width=\linewidth]{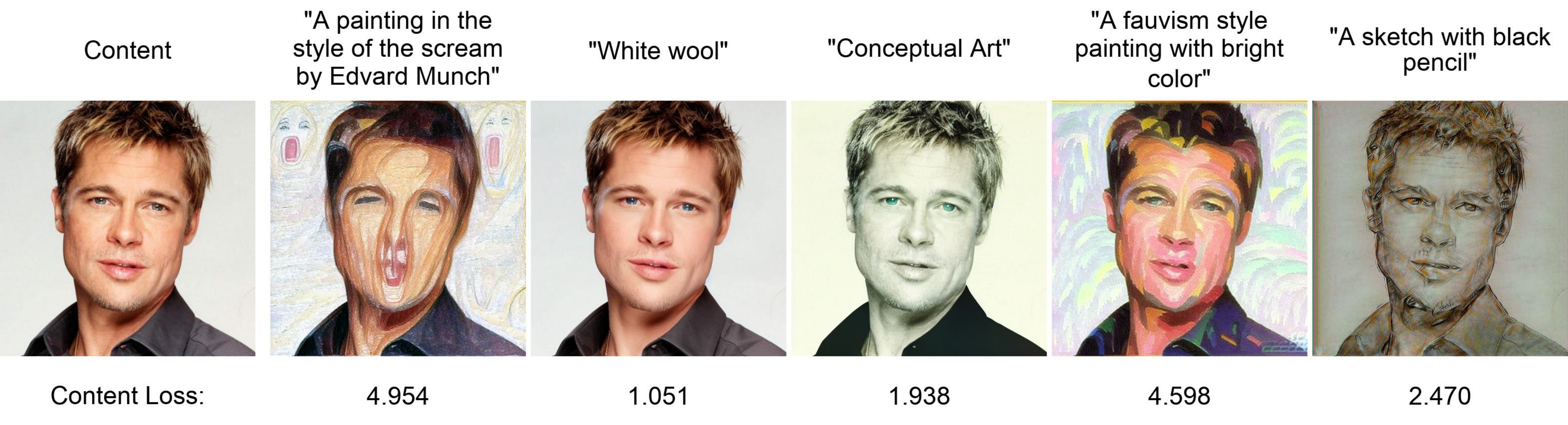}
    \caption{\textbf{Problem of the content loss.} This figure shows randomly picked stylized images and their content loss values. Note that they are obtained from \textbf{different randomly selected style transfer methods}. The method names are intentionally hidden to ensure unbiased perception. The fisrt image is the original content image and the remaining ones are the stylized images. Style text descriptions are shown above the images. Content loss values are shown below the images. The nearly reconstructed stylized image (2nd) consistently achieves the best content loss. More interestingly, the well-stylized image (4th) has a inferior content loss nearly identical to the completely distorted image (1st). This indicates that content loss struggles to differentiate between style variations (4th) and content distortions (1st).
    }
\label{fig:content_loss_2}
\end{figure*}

\begin{figure*}
    \centering
    \includegraphics[width=\linewidth]{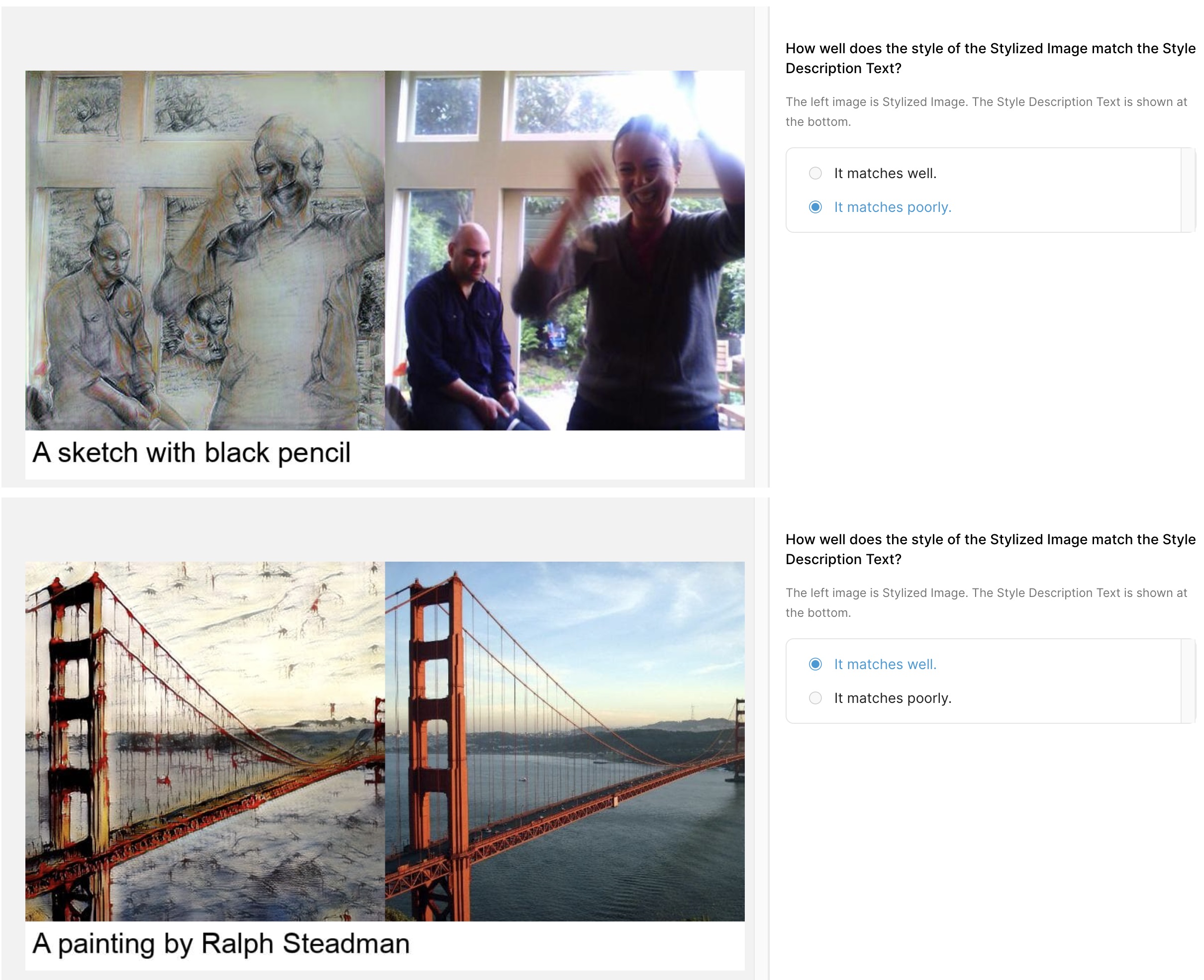}
    \caption{\textbf{User interface for image evaluation in user study.} Here we evaluate \textit{Style Consistency}.}
    \label{fig:supp_anno_style}
\end{figure*}

\begin{figure*}
    \centering
    \includegraphics[width=\linewidth]{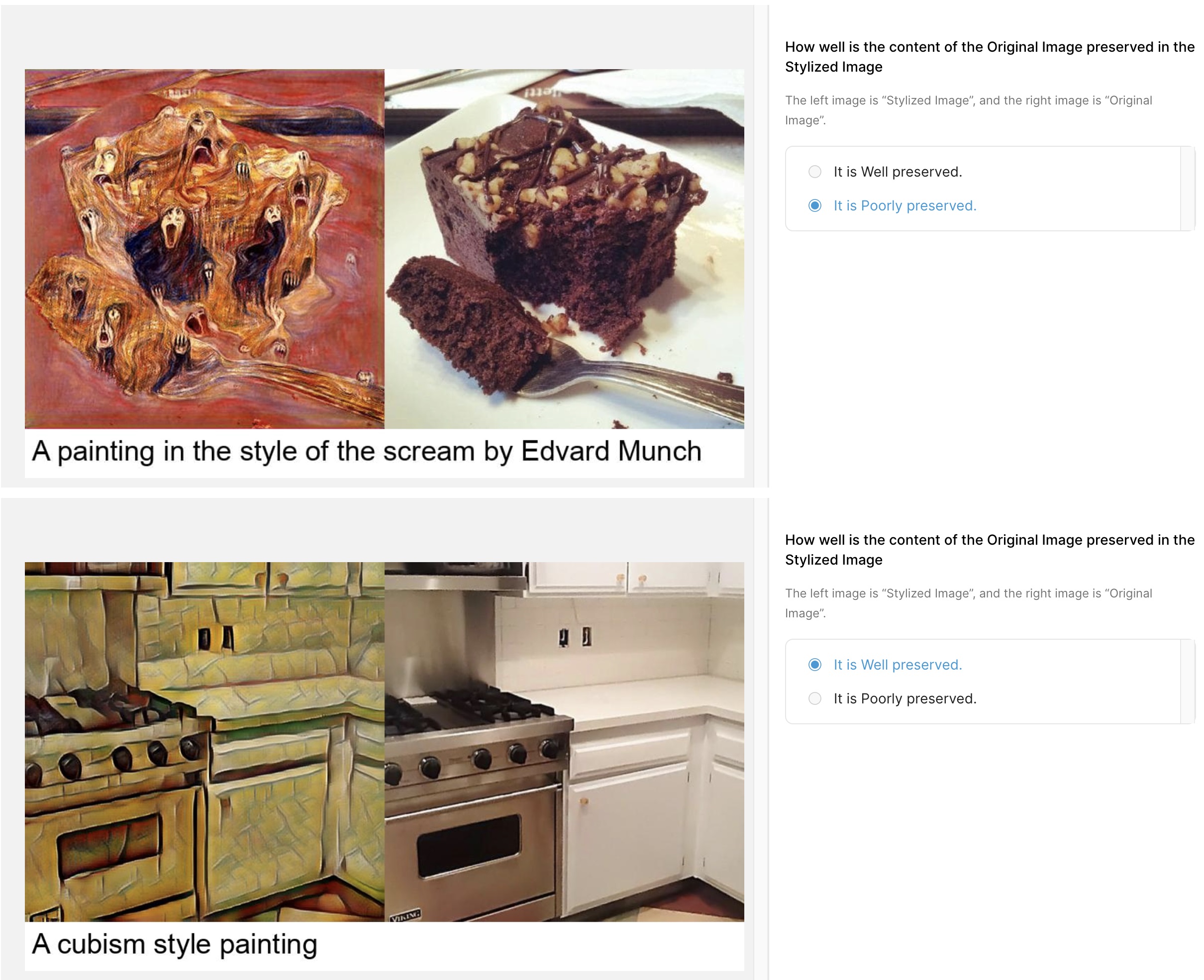}
    \caption{\textbf{User interface for image evaluation in user study.} Here we evaluate \textit{Content Preservation}.}
    \label{fig:supp_anno_content}
\end{figure*}

\begin{figure*}
    \centering
    \includegraphics[width=\linewidth]{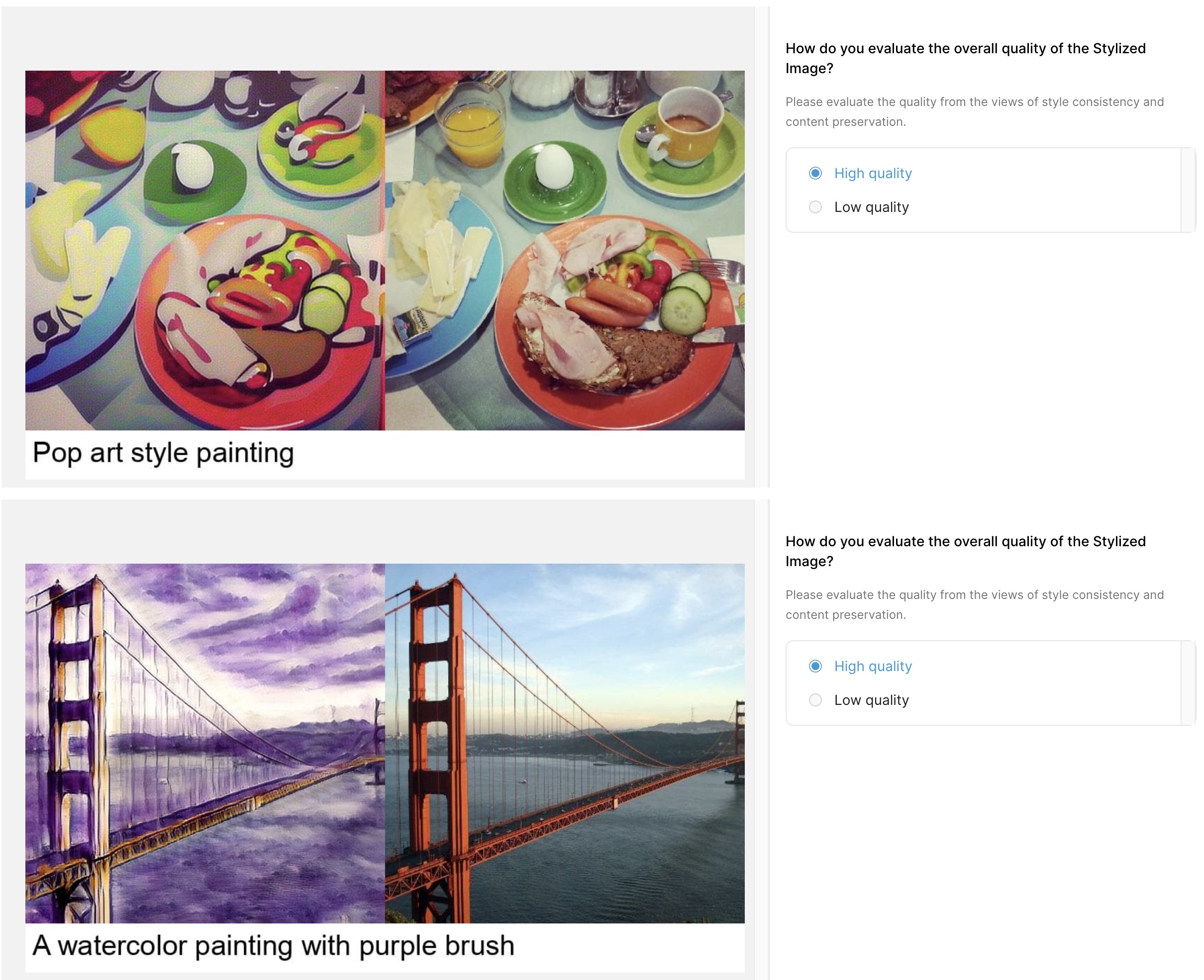}
    \caption{\textbf{User interface for image evaluation in user study.} Here we evaluate \textit{Overall Quality}.}
    \label{fig:supp_anno_overall}
\end{figure*}

\begin{figure*}
    \centering
    \includegraphics[width=\linewidth]{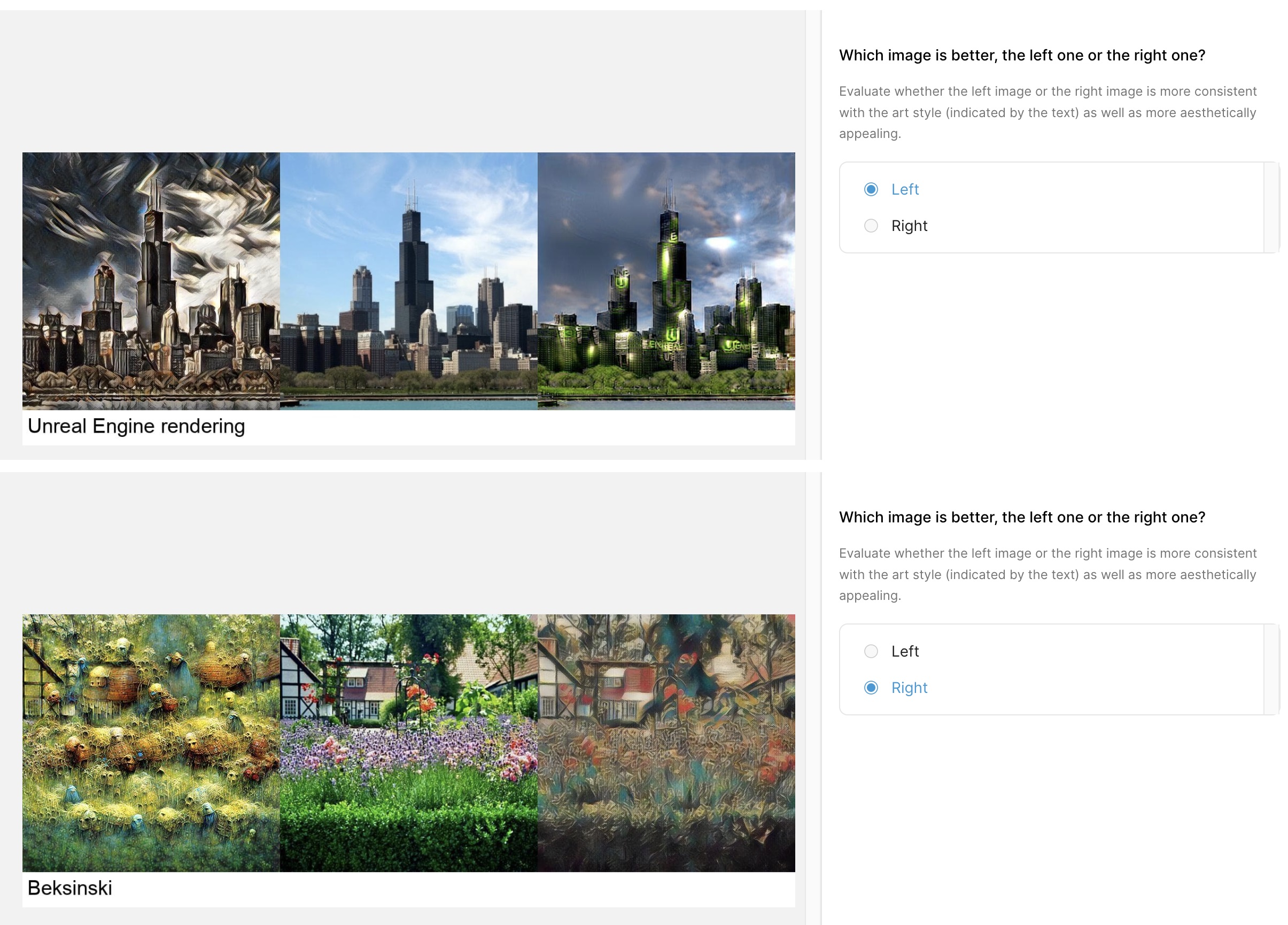}
    \caption{\textbf{User interface for image evaluation in user study.} This user interface is used for ablation studies.}
    \label{fig:supp_anno_ab}
\end{figure*}

\begin{figure*}
    \centering
    \includegraphics[width=\linewidth]{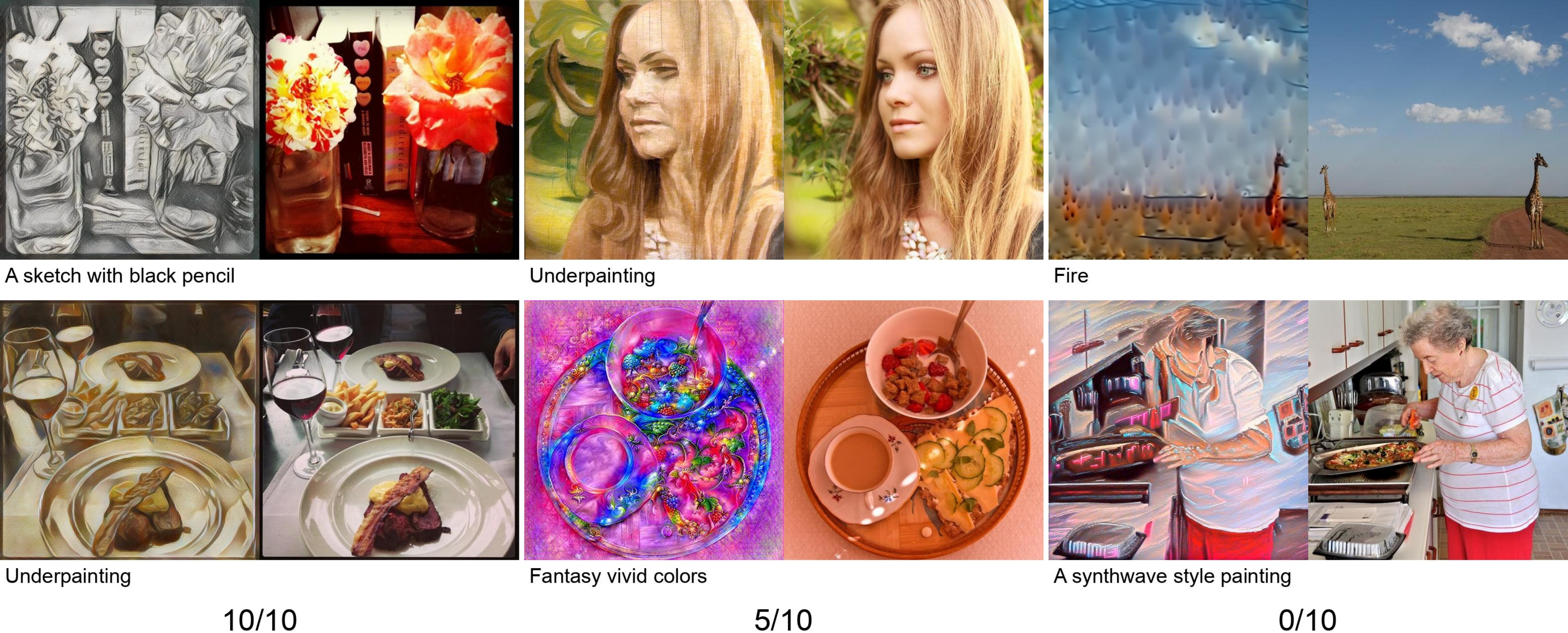}
    \caption{\textbf{Examples of \textit{Style Consistency} annotations.} At the bottom of each column we show the number of positive responses received over the total response number.}
    \label{fig:supp_anno_eg_style}
\end{figure*}

\begin{figure*}
    \centering
    \includegraphics[width=\linewidth]{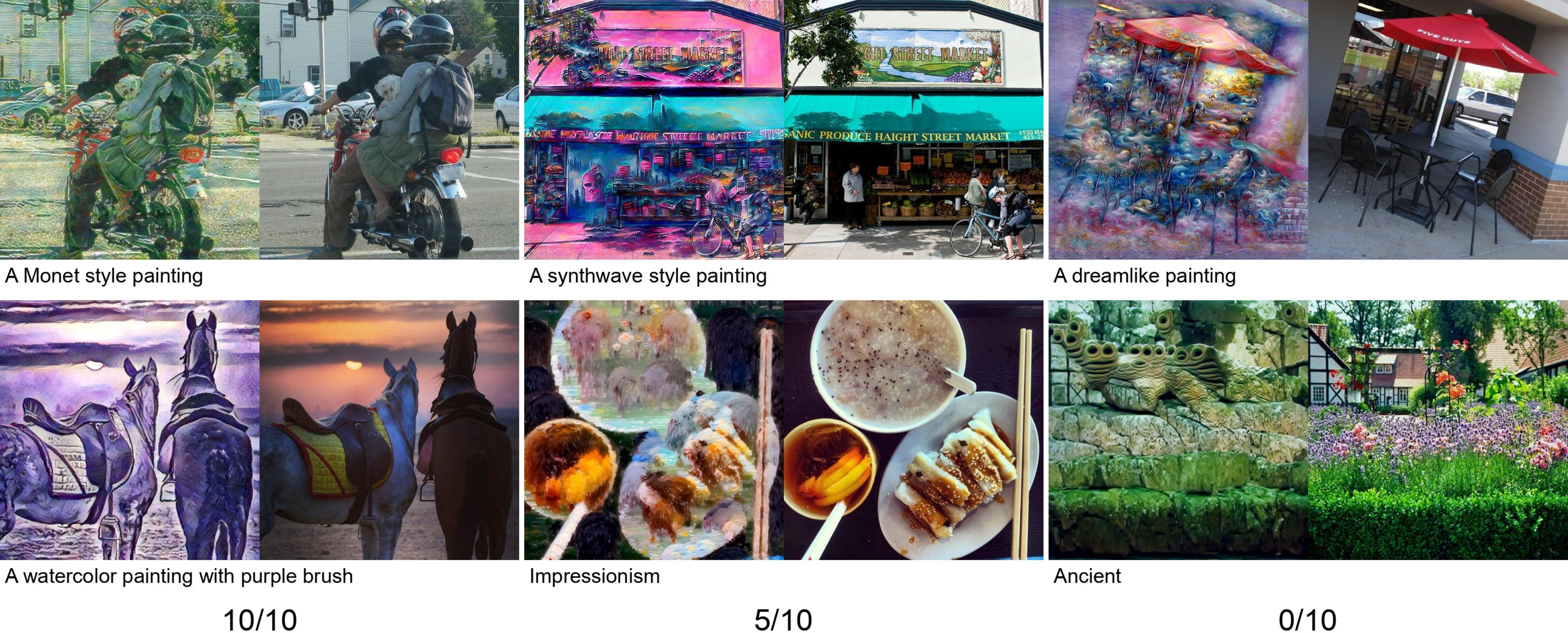}
    \caption{\textbf{Examples of \textit{Content Preservation} annotations.} At the bottom of each column we show the number of positive responses received over the total response number.}
    \label{fig:supp_anno_eg_content}
\end{figure*}

\begin{figure*}
    \centering
    \includegraphics[width=\linewidth]{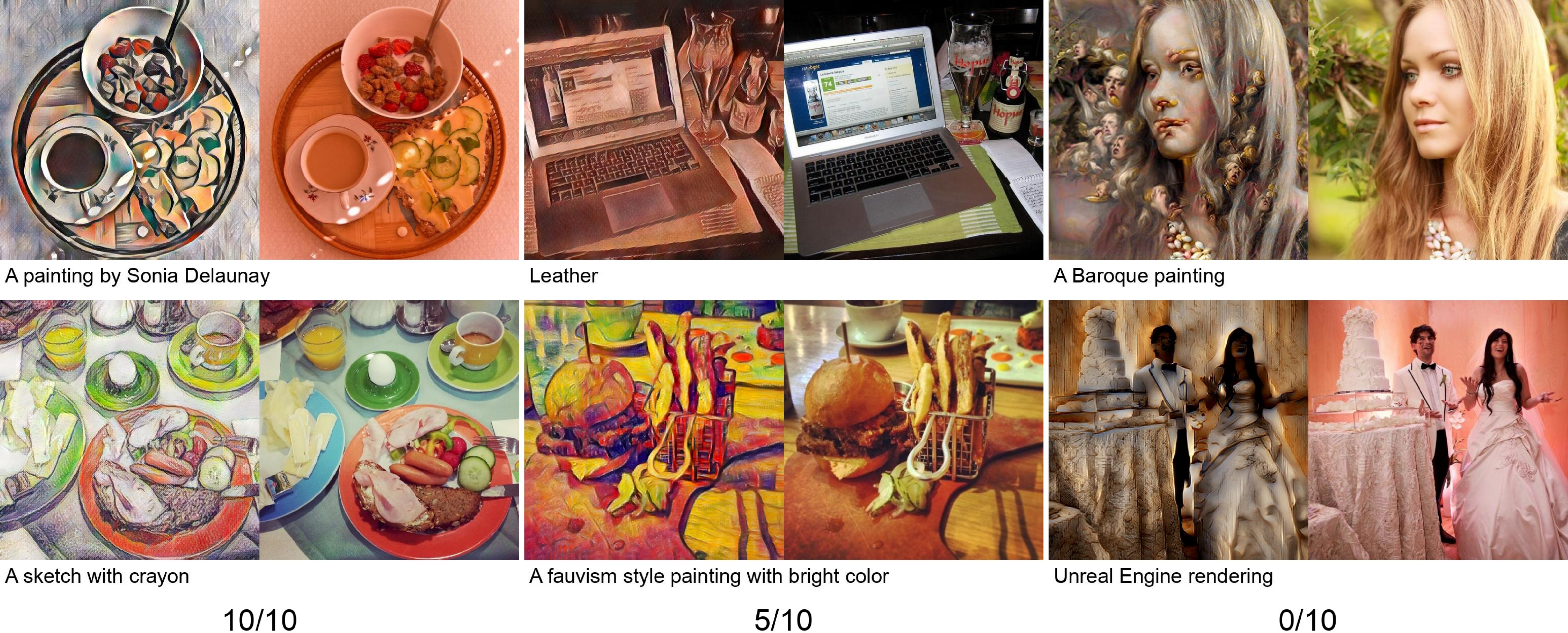}
    \caption{\textbf{Examples of \textit{Overall Quality} annotations.} At the bottom of each column we show the number of positive responses received over the total response number.}
    \label{fig:supp_anno_eg_overall}
\end{figure*}

\begin{table}[t]
    \centering
    \caption{\textbf{Number of annotators involved in each evaluation task in our user study.} \textit{Style}, \textit{Content}, and \textit{Overall} are the three aspects in our main user study. \textit{Ablation} refers to all ablation experiments.}
    \begin{tabular}{@{}l|c@{}}
        \toprule
        Task & Number of Annotators \\
        \midrule
        Style & 872 \\
        Content & 810 \\
        Overall & 939 \\
        Ablation & 5041 \\
        \bottomrule
    \end{tabular}
    \label{tab:user_study_design}
    \vspace{-0.2cm}
\end{table}

\begin{table}[t]
    \centering
    \caption{\textbf{Additional ablation study on different design choices.} The performance is evaluated through user study. For all these design choices, the user preference percentages are less than 50\%, indicating that they are inferior to our method in the main paper.}
    \begin{tabular}{@{}lc@{}}
        \toprule
        Setting & Preference \% $\uparrow$ \\
        \midrule
        Retrieval + AdaAttN & 31.1  \\
        StableDiffusion + AdaAttN & 35.5  \\
        ExcludeInv4 & 44.5  \\         
        ExcludeInv8 & 44.9  \\   
        + GlobalLoss & 49.3  \\   
        \bottomrule
    \end{tabular}
    \label{tab:supp_ab_user_study}
\end{table}

\section{Style Aggregation Strategy}
In Multi-style Boosting (Section~5.1 of the main paper), we propose to aggregate styles $\sr$ to enhance style transfer quality. The aggregation strategy depends on the specific implementation of the IIST method $\adaattn$. Here we briefly describe the straightforward aggregation algorithm for AdaAttN~\cite{liu2021adaattn} as an example of $\adaattn$. Similar to many IIST model~\cite{huang2017arbitrary, park2019arbitrary}, AdaAttN $\adaattn$ can be decomposed into a feature extraction network $\adaattn_f$ and a style transfer module $\adaattn_t$. Attention mechanism is used for $\adaattn_t$ to process the output features from $\adaattn_f$ in AdaAttN.

After obtaining $\sr$ from cross-modal GAN inversion, we feed them into the feature extraction network $\adaattn_f$ separately and concatenate the outputs together over the sequential dimension at the attention layers in $\adaattn_t$. Since attention layers adaptively focus on the best-matching regions, they can benefit significantly from the high-quality style patterns in the concatenated style representations, while being free from the negative impact of low-quality patterns. The concatenated feature is the $F$ in the Algorithm~2 of the main paper, which is directly used by $\adaattn_t$ to apply style transfer.

\section{Latent Initialization}

Similar to traditional GAN inversion~\cite{abdal2019image2stylegan}, in cross-modal GAN inversion, the quality of the generated image is sensitive to the initial value of $w$. 
Traditional GAN inversion methods often choose the mean latent $\Bar{w}$ of the dataset as the initial $w$. Unfortunately, this initialization is not suitable for our problem as there is no style text description dataset available to compute $\Bar{w}$. To overcome this issue, we propose to randomly sample a set of $w$, from which we pick the best one based on Eq.~3 in the main paper. Formally, we first sample multiple $z_i \sim \mathcal{N}(0,1)$. Then we run the mapping network of StyleGAN3 on them to obtain $\{w_i\}$. Finally, we calculate
\begin{align}
\begin{split}
    \label{eq:init}
    \hat{w} = \argmin_{w \in \{w_i\}} L_\text{sty},
\end{split}
\end{align}
as the initial value of the StyleGAN3 latent embedding.

\section{Additional Ablation Study}
\subsection{Qualitative Ablation Study}

In order to visually explore the impact of various design choices, we conduct a qualitative ablation study illustrated in \cref{fig:supp_ab_qual}. All of the design choices outlined in Table 4 of the main paper are considered. The results indicate that a crop size of 128 (CropSize128) often leads to either over-stylization or under-stylization. Furthermore, the effect of different patch loss weights (PatchLoss500, PatchLoss2500) is negligible, which aligns with the user preference data presented in Table 4 of the main paper. While omitting patch augmentation (NoAug) typically has a minimal effect on the quality of the stylized images, it can sometimes lead to errors such as the incorrect highlighting of edges in the stylized image shown in the first row. In contrast, the omission of patch cropping (NoCrop) can have a more pronounced effect, resulting in oversimplified styles. Finally, our ablation study confirms the importance of multi-style boosting, as performance is significantly degraded when it is not utilized (NoBoosting).

\subsection{Additional User Study}
We report user study results for additional ablation study. We follow the settings of the ablation user study conducted in our main paper, and consider the text-guided image style transfer task.  We report the user preference percentage for the additional 
design choices in Table~\ref{tab:supp_ab_user_study}. 

Specifically, we first consider replacing our cross-modal GAN inversion by image retrieval and a text-to-image generative model, respectively, \ie, Retrieval + AdaAttN and StableDiffusion + AdaAttN. For image retrieval, we use CLIP image embedding to retrieve a style representation from WikiArt  dataset, which is the same dataset that the StyleGAN3 model was trained on. For the text-to-image generative model, we use the open-source implementation StableDiffusion\footnote{\url{https://github.com/CompVis/stable-diffusion}} of the LDMs~\cite{rombach2021highresolution}. We observe that our method significantly outperforms these two design choices, demonstrating the effectiveness of our cross-modal GAN inversion method even if only the text-guided image style transfer task is considered. 

Next, we explore if the entire $\mathcal{W}^+$ space is important to ensure the style transfer quality. Inspired by \cite{xia2021tedigan, xia2021towards}, we consider excluding the first 4 layers or 8 layers from the inversion, \ie, ExcludeInv4 and ExcludeInv8. However, we observe that these partial inversion techniques have a negative impact on the style transfer quality.

Finally, inspired by \cite{kwon2022clipstyler}, we consider adding a global CLIP loss to the objective function of our cross-modal GAN inversion, \ie, a CLIP loss without image patch cropping. User study result indicates that this additional loss does not improve the user preference percentage. Therefore, we do not add this loss to our main method.

\section{Additional Qualitative Results}

\noindent\textbf{Inverted Style Representation Examples.} \cref{fig:inv_examples} shows some examples of the inverted style representations from style text descriptions. We can observe that many of them do not contain meaningful content, however, they all exhibit certain styles corresponding to the input style text descriptions.

\noindent\textbf{Additional Comparisons with TIST Methods.} We show comparison results on more text-image pairs in Figure~\ref{fig:supp_cmp}. These examples consistently demonstrate the overall superiority of our method.

\noindent\textbf{Additional MMIST Results.}
We show more multimodality-guided image style transfer results in Firgure~\ref{fig:supp_multi}. These examples demonstrate how our method combines different styles and faithfully applies them to various content images. 

\newpage
\newpage

\noindent\textbf{Additional Results of MMIST with Four Style Sources and Cross-modal Style Interpolation.} We also show additional MMIST resutls with style interpolation in Figures~\ref{fig:supp_four_1} and \ref{fig:supp_four_2}. Same as Figure 6 in our main paper, these figures show style interpolation results between 2 text style descriptions and 2 style images. The interpolation ratio for each column or row is fixed to be 1:0,\ \  0.75:0.25,\ \  0.5:0.5,\ \  0.25:0.75,\ \  0:1.

\section{Ineffectiveness of Content Loss}

Initially, we considered to utilize the content loss employed by CLIPStyler~\cite{kwon2022clipstyler} as a metric for quantitative evaluation. However, both our theoretical insights and practical experiments indicated that this content loss doesn not align with human perception. 

The content loss as defined in CLIPStyler~\cite{kwon2022clipstyler} is calculated as the MSE Loss between the deep VGG features of the stylized image and the content image. Given that VGG is pretrained for recognition tasks, its deep features are acutely sensitive to the distinct visual cues of an input image, such as color and texture. However, variations in color and texture do not necessarily correlate with alterations in the content as perceived by humans. Moreover, modifications in color and texture are the essential outcomes of the style transfer process. This implies that a smaller content loss might indicate a less effective stylization outcome. In the extreme case, an identity mapping function preserves all the content information and has the smallest content loss, but it is a trivial style transfer process and thus undesired.
Therefore, while employing content loss during training is not problematic due to the concurrent use of style loss, which ensures the style quality, its application as an evaluation metric is unsuitable.

To further validate our analysis regarding the limitations of the content loss defined in CLIPStyler~\cite{kwon2022clipstyler}, we randomly pick stylized images from different style transfer methods and compute their content loss for comparison. The results are shown in Figures~\ref{fig:content_loss_1} and \ref{fig:content_loss_2}. Note that the names of selected methods are intentionally hidden to ensure unbiased perception. In Figure~\ref{fig:content_loss_1}, the highly stylized images (2nd and 4th) appear to incur a higher content loss, even though they largely preserve the original content. In contrast, the 5th stylized image, which deviates minimally from the content image, achieves the best content loss. In Figure~\ref{fig:content_loss_2}, the stylized image (2nd) which nearly reconstructs the original image consistently achieves the best content loss. More interestingly, the well-stylized image (4th) has a inferior content loss nearly identical to the completely distorted image (1st). This indicates that content loss struggles to differentiate between style variations (4th) and content distortions (1st). In summary, the content loss defined in CLIPStyler appears ill-equipped to differentiate between style modifications and content distortions. Therefore, we opt not to use content loss as an evaluation metric in this paper.

\section{Limitation}

While our approach proves robust across various applications, it is intrinsically constrained by its reliance on the pretrained style representation generator, the adapted IIST method, and the CLIP model, which is utilized to construct the loss function in the cross-modal GAN inversion algorithm.

We utilize the WikiArt pretrained StyleGAN3 as our style representation generator. While this model encompasses a broad spectrum of styles, its effectiveness can be compromised when confronted with out-of-distribution styles. This limitation arises from the finite scope of the WikiArt dataset. Consequently, when presented with certain styles that are outside this dataset's domain, our style transfer outcomes might not achieve the desired quality.

Similarly, the efficacy of our solution is significantly influenced by the adapted IIST method. This component executes the style transfer after the generation of intermediate style representations. If the adapted IIST method manifests any limitations or biases, it can have a direct negative impact on the results generated by our method.

Moreover, the CLIP model and the Style-specific CLIP Loss are not perfect. Potential inaccuracies in these parts may yield imprecise intermediate style representations, further influencing the quality of the stylized images.

In addition, our method requires a per-sytle optimization procedure for fast style transfer. However, this optimization can be time-intensive, potentially hindering our method's application in time-sensitive scenarios. An alternative could be training a feed-forward style transfer network to eliminate the need for per-style optimization.  We leave this potential improvement direction as future work.

\clearpage

\begin{figure*}[t]
    \centering
    \includegraphics[width=0.9\linewidth]{figures/supp/supp_comparisons.pdf}
    \caption{\textbf{Additional comparison with other TIST methods.}}
    \label{fig:supp_cmp}
\end{figure*}

\begin{figure*}[t]
    \centering
    \includegraphics[width=0.9\linewidth]{figures/supp/supp_multi_results.pdf}
    \caption{\textbf{Additional MMIST results.}}
    \label{fig:supp_multi}
\end{figure*}

\begin{figure*}[t]
    \centering
    \includegraphics[width=\linewidth]{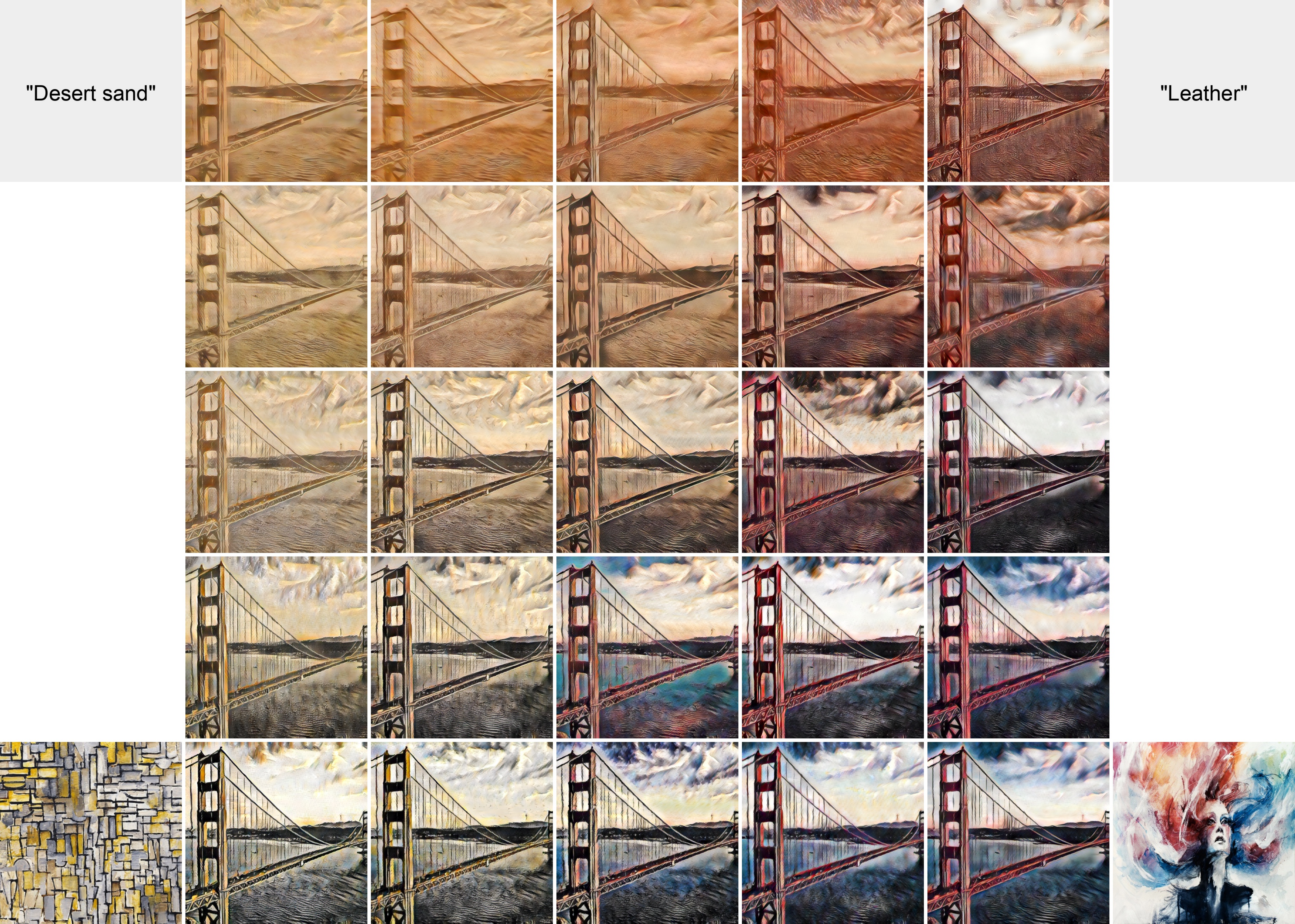}
    \caption{\textbf{Additional MMIST results with four image and text styles and style interpolation.} (1)}
    \label{fig:supp_four_1}
\end{figure*}

\begin{figure*}[t]
    \centering
    \includegraphics[width=\linewidth]{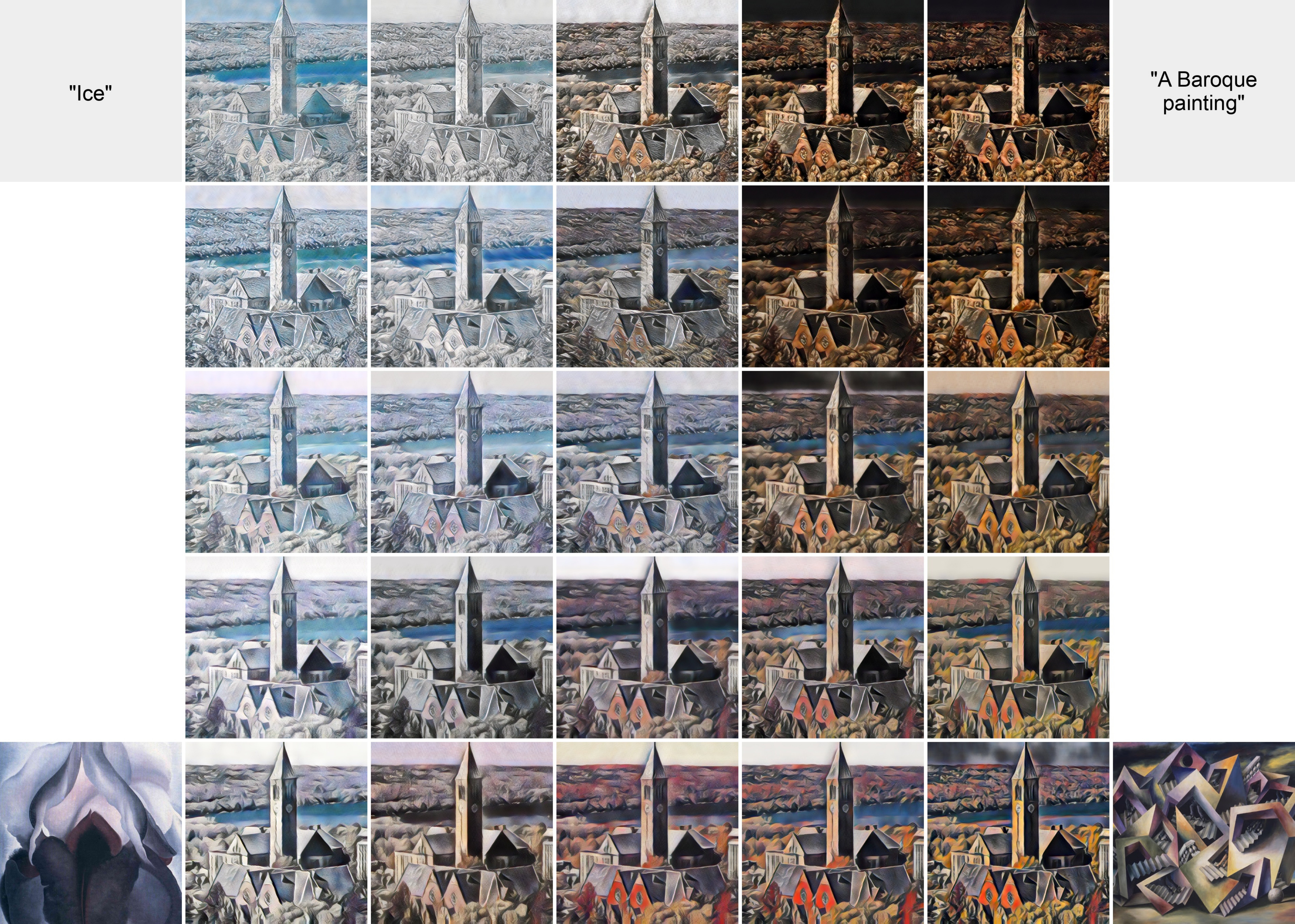}
    \caption{\textbf{Additional MMIST results with four image and text styles and style interpolation.} (2)}
    \label{fig:supp_four_2}
\end{figure*}

\end{document}